%% file: VFOA.tex
\documentclass[10pt,twocolumn,twoside]{IEEEtran}
\input{header}
\input{notations}
\begin{document}

\input{title}
\input{section_abstract}

%\section{Introduction}
\input{section_introduction}

%\section{Related Works}
\input{section_related_works}

%\section{Model}
\input{section_model}
\input{section_inference}

%\section{Learning}
\input{section_learning}

%\section{Experiments}
\input{section_implementation}
\input{section_experimental_results}
\input{section_conclusion}

\appendices
%% \section{VFOA Transition Probabilities}
%% \label{app:transitions_list}
%% Please consult the supplemental material.
%% \section{VFOA Learning}
%% \label{app:vfoa-learning}
%% Please consult the supplemental material.

\input{appendix_vfoa_transition_cases}

\input{appendix_vfoa_learning}

\section*{Acknowledgments}

The authors would like to thank Vincent Drouard for his valuable expertise in head pose estimation and tracking.

\ifCLASSOPTIONcaptionsoff
  \newpage
\fi

\bibliographystyle{IEEEtran}
% \bibliography{biblio}


\begin{IEEEbiography}[{\includegraphics[width=1in,height=1.25in,clip,keepaspectratio]{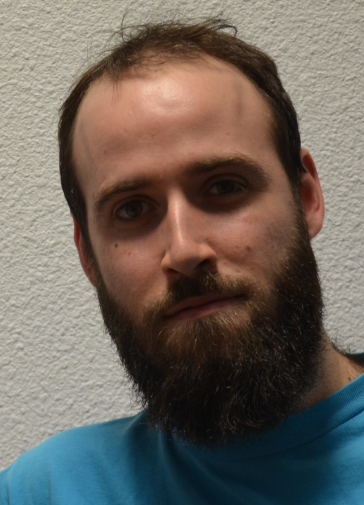}}]{Benoit Mass\'e}
received the M.Eng. degree in applied mathematics and computer science from  ENSIMAG, Institut National Polytechnique de Grenoble, France, in  2013, and the M.Sc. degree in graphics, vision and robotics from Universit\'e Joseph Fourier, Grenoble, France, in 2014.
Currently he is a PhD student in the PERCEPTION team at INRIA Grenoble Rhone-Alpes. His research interests include scene understanding, machine learning and computer vision, with special emphasis on attention recognition for human-robot interaction.
\end{IEEEbiography}

\begin{IEEEbiography}[{\includegraphics[width=1in,height=1.25in,clip,keepaspectratio]{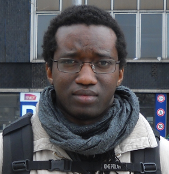}}]{Sil\`{e}ye Ba}
received the M.Sc. (2000) in applied mathematics and signal processing from University of Dakar, Dakar, Senegal, and the M.Sc. (2002) in mathematics, computer vision, and machine learning from Ecole Normale Sup\'erieure de Cachan, Paris, France. From
2003 to 2009 he was a PhD student and then a post-doctoral researcher at IDIAP Research Institute, Martigny, Switzerland,
where he worked on probabilistic models for object tracking and human activity recognition. From 2009 to 2013, he was a researcher at Telecom Bretagne, Brest, France working on variational models for multi-modal geophysical data processing. From 2013 to 2014 he worked at RN3D Innovation Lab, Marseille, France, as a research engineer, where he used computer vision and machine learning principles and methods to develop human-computer interaction software tools. From  2014 to 2016 he was a researcher in the PERCEPTION
team at INRIA Grenoble Rh\^one-Alpes, working on machine learning and computer vision models for human-robot interaction. Since May 2016 he is a computer vision scientist with VideoStitch, Paris.
\end{IEEEbiography}

\begin{IEEEbiography}[{\includegraphics[width=1in,height=1.25in,clip,keepaspectratio]{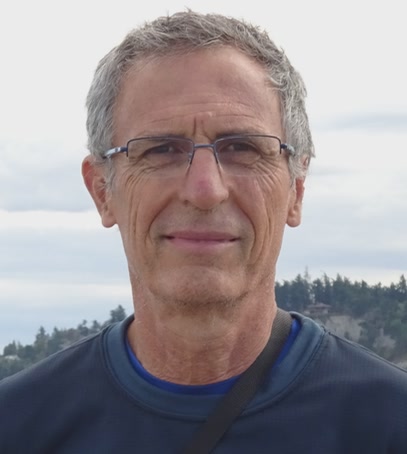}}]{Radu Horaud}
received the B.Sc. degree in Electrical Engineering, the M.Sc. degree
in Control Engineering, and the Ph.D. degree in Computer Science from
the Institut National Polytechnique de Grenoble, France. In 1982-1984 he was a post-doctoral fellow with the Artificial Intelligence Center, SRI International, Menlo Park, CA.
Currently he holds a position of director of research with INRIA Grenoble Rh\^one-Alpes, where
he is the founder and head of the PERCEPTION team. His
research interests include computer vision, machine learning, audio signal processing,
audiovisual analysis, and robotics. Radu Horaud and his collaborators received numerous best paper awards. He is an area editor of the
\textit{Elsevier Computer Vision and Image Understanding}, a member of
the advisory board of the \textit{Sage International Journal of Robotics
  Research}, and an associate editor of the
\textit{Kluwer International Journal of Computer Vision}.
He was program co-chair of IEEE ICCV'01 and of ACM ICMI'15.
In 2013, Radu Horaud was awarded an ERC Advanced Grant for his project \textit{Vision and Hearing in Action} (VHIA) and in 2017 he was awarded an ERC Proof of Concept Grant for this project VHIALab.
\end{IEEEbiography}

\end{document}

%% file: header.tex
\ifCLASSINFOpdf
  \usepackage[pdftex]{graphicx}
  \DeclareGraphicsExtensions{.png}
\else
\fi
\usepackage[english]{babel}
\usepackage[T1]{fontenc}
\usepackage[utf8]{inputenc}
\usepackage{amsmath}
\usepackage{amssymb}
\usepackage{gensymb}
\usepackage{amsfonts}
\usepackage{mathtools}
\usepackage{listings}
\usepackage{dsfont}
\usepackage{color}
\usepackage{textcomp}
\usepackage{url}
\usepackage{tikz}
\usetikzlibrary{arrows}

\usepackage{algorithm}
\usepackage[noend]{algpseudocode}

\makeatletter
\def\BState{\State\hskip-\ALG@thistlm}
\newsavebox\myboxA
\newsavebox\myboxB
\newlength\mylenA
\newcommand*\xoverline[2][0.75]{%
    \sbox{\myboxA}{$\m@th#2$}%
    \setbox\myboxB\null% Phantom box
    \ht\myboxB=\ht\myboxA%
    \dp\myboxB=\dp\myboxA%
    \wd\myboxB=#1\wd\myboxA% Scale phantom
    \sbox\myboxB{$\m@th\overline{\copy\myboxB}$}%  Overlined phantom
    \setlength\mylenA{\the\wd\myboxA}%   calc width diff
    \addtolength\mylenA{-\the\wd\myboxB}%
    \ifdim\wd\myboxB<\wd\myboxA%
       \rlap{\hskip 0.5\mylenA\usebox\myboxB}{\usebox\myboxA}%
    \else
        \hskip -0.5\mylenA\rlap{\usebox\myboxA}{\hskip 0.5\mylenA\usebox\myboxB}%
    \fi}
\def\ignorecitefornumbering#1{%
    \begingroup
        \@fileswfalse
         #1%                     % do \cite comand
    \endgroup
}
\makeatother

\def\pizz  {p_{1}}
\def\pijz  {p_{2}}
\def\pizk  {p_{3}}
\def\pikk  {p_{4}}
\def\pijk  {p_{5}}
\def\pizkz {p_{6}}
\def\pikkz {p_{7}}
\def\pijkz {p_{8}}
\def\pizki {p_{9}}
\def\pikki {p_{10}}
\def\pijki {p_{11}}
\def\pizkl {p_{12}}
\def\pikkl {p_{13}}
\def\pilkl {p_{14}}
\def\pijkl {p_{15}}

\def\pizzhat  {\hat{p}_{1}}
\def\pijzhat  {\hat{p}_{2}}
\def\pizkhat  {\hat{p}_{3}}
\def\pikkhat  {\hat{p}_{4}}
\def\pijkhat  {\hat{p}_{5}}
\def\pizkzhat {\hat{p}_{6}}
\def\pikkzhat {\hat{p}_{7}}
\def\pijkzhat {\hat{p}_{8}}
\def\pizkihat {\hat{p}_{9}}
\def\pikkihat {\hat{p}_{10}}
\def\pijkihat {\hat{p}_{11}}
\def\pizklhat {\hat{p}_{12}}
\def\pikklhat {\hat{p}_{13}}
\def\pilklhat {\hat{p}_{14}}
\def\pijklhat {\hat{p}_{15}}

%% file: notations.tex
\def\X  {\mathbf{X}}
\def\V  {\mathbf{V}}
\def\VV {\mathrm{V}}
\def\G  {\mathbf{G}}
\def\Gdot  {\mathbf{\dot G}}
\def\R  {\mathbf{R}}
\def\Rdot  {\mathbf{\dot R}}
\def\H  {\mathbf{H}}
\def\LL {\mathbf{L}}

\def\I       {\mathbf{I}}
\def\J       {\mathbf{J}}
\def\KK      {\mathbf{K}}
\def\PP      {\mathbf{P}}
\def\A       {\mathbf{A}}

\def\C       {\mathbf{C}}

\def\bb      {\mathbf{b}}
\def\zero    {\mathbf{0}}
\def\Sigmabf {\boldsymbol{\Sigma}}
\def\mubf    {\boldsymbol{\mu}}
\def\Gammabf {\boldsymbol{\Gamma}}

\def\alphabf {\boldsymbol{\alpha}}
\def\betabf  {\boldsymbol{\beta}}

\def\thetabf {\boldsymbol{\theta}}

\def\eg{\emph{e.g}. } 
\def\ie{\emph{i.e}. }

\def\wrt{w.r.t. }

\DeclareMathOperator*{\argmax}{argmax}
\DeclareMathOperator{\Tr}{Tr}
\DeclareMathOperator{\diag}{Diag\,}

%% file: title.tex
\title{Tracking Gaze and Visual Focus of Attention of People Involved in Social Interaction}

% \author{Michael~Shell,~\IEEEmembership{Member,~IEEE,}
%         John~Doe,~\IEEEmembership{Fellow,~OSA,}
%         and~Jane~Doe,~\IEEEmembership{Life~Fellow,~IEEE}% <-this % stops a space
% \thanks{M. Shell was with the Department
% of Electrical and Computer Engineering, Georgia Institute of Technology, Atlanta,
% GA, 30332 USA e-mail: (see http://www.michaelshell.org/contact.html).}% <-this % stops a space
% \thanks{J. Doe and J. Doe are with Anonymous University.}% <-this % stops a space
% \thanks{Manuscript received April 19, 2005; revised August 26, 2015.}}

\author{Benoit~Mass\'e,~%\IEEEmembership{Member,~IEEE,}
        Sil\`eye~Ba,~%\IEEEmembership{Fellow,~OSA,}
        and~Radu~Horaud%,~\IEEEmembership{Life~Fellow,~IEEE}% <-this % stops a space
\thanks{B. Mass\'e and R. Horaud are with INRIA Grenoble Rh\^one-Alpes, Montbonnot Saint-Martin, France.}% <-this % stops a space
\thanks{S. Ba is with Dailymotion, Paris, France}% <-this % stops a space
\thanks{This work is supported by ERC Advanced Grant VHIA \#340113.}}

\maketitle

%% file: section_abstract.tex
\begin{abstract}
The visual focus of attention (VFOA) has been recognized as a prominent conversational cue. We are interested in estimating and tracking the VFOAs associated with multi-party social interactions. We note that in this type of situations the participants either look at each other or at an object of interest; therefore their eyes are not always visible. Consequently both gaze and VFOA estimation cannot be based on eye detection and tracking. We propose a method that exploits the correlation between eye gaze and head movements. Both VFOA and gaze are modeled as latent variables in a Bayesian switching state-space model. The proposed formulation leads to a tractable learning procedure and to an efficient algorithm that simultaneously tracks gaze and visual focus. The method is tested and benchmarked using two publicly available datasets that contain typical multi-party human-robot and human-human interactions.

%We address the problem of estimating the visual focus of attention (VFOA), \eg who is looking at whom?
%This is of particular interest in human-robot interactive scenarios, \eg when the task requires to identify targets of interest over time. The paper makes the following contributions. We propose a Bayesian temporal model that connects VFOA to gaze direction and to head pose. Model inference is then cast into a switching Kalman filter formulation, which makes it tractable. The model parameters are estimated via training based on manual annotations. 
%
%The method is tested and benchmarked using a publicly available dataset. We show that both the gaze and the VFOA of several persons can be reliably and simultaneously estimated over time from observed head poses as well as from people and object locations. On average, our method compares favorably with two other methods.
\end{abstract}

\begin{IEEEkeywords}
Visual focus of attention, eye gaze, head pose, dynamic Bayesian model, switching Kalman filter, multi-party dialog, human-robot interaction.
\end{IEEEkeywords}

% \fi
\IEEEpeerreviewmaketitle

%% file: section_introduction.tex
\section{Introduction}
\label{sec:introduction}
\IEEEPARstart{I}{}n this paper we are interested in the computational analysis of social interactions. In addition to speech, people communicate via a large variety of non-verbal cues, \eg prosody, hand gestures, body movements, head nodding, eye gaze, and facial expressions. For example, in a \textit{multi-party} conversation, a common behavior consists in looking either at a person, \eg the speaker, or at an object of current interest, \eg a computer screen, a painting on a wall, or an object lying on a table.
We are particularly interested in estimating the \textit{visual focus of attention} (VFOA), or who is looking at whom or at what, which has been recognized as one of the most prominent social cues. It is used in multi-party dialog to establish face-to-face communication, to respect social etiquette, to attract someone's attention, or to signify speech-turn taking, thus complementing speech communication.

%In human-robot interaction (HRI), the robot must be able to keep temporal knowledge about the locations of the participants, to discriminate between the speaker and the listeners, to recognize the objects of interest, etc.
%Moreover, in addition to spoken dialog management, it is desirable that a robot exhibits human-like behavior, appropriate for the situation at hand, \eg head nodding, hand gestures, etc.

%
%Among these cues, visual focus of attention (VFOA) estimation of multiple persons provides answers to: \textit{Who is looking at whom?} \textit{Who is looking at what?} \textit{Who is the speaker?} \textit{Who are the listeners?} \etc
%The joint multi-person VFOA analysis provides insights on who is the current speaker or which is the object of interest.

%%% Term definition
The VFOA characterizes a perceiver/target pair. It may be defined either by the line from the perceiver's face to the perceived target, or by the perceiver's \textit{direction of sight} or \textit{gaze direction} (which is often referred to as eye gaze or simply gaze).  Indeed, one may state that the VFOA of person $i$ is target $j$ if the perceiver's gaze is aligned with the perceiver-to-target line.
From a physiological point of view, eye gaze depends on both eyeball orientation and head orientation. Both the eye and the head are rigid bodies with three and six degrees of freedom respectively. The head position (three coordinates) and the head orientation (three angles) are jointly referred to as the \textit{head pose}. With proper choices for the head- and eye-centered coordinate frames, one can assume that gaze is a combination of head pose and of eyeball orientation,\footnote{Note that orientation generally refers to the pan, tilt and roll angles of a rigid-body pose, while direction refers to the polar and azimuth angles or, equivalently, a unit vector. Since the contribution of the roll angle to gaze is generally marginal, in this paper we make no distinction between orientation and direction.} and the VFOA depends on head pose, eyeball orientation, and target location.

In this paper we are interested into estimating and tracking jointly the VFOAs of a group of people that communicate with each other and with a robot, or \textit{multi-party} HRI (human-robot interaction), which may well be viewed as a generalization of \textit{single-user} HRI. From a methodological point of view the former is more complex than the latter. Indeed, in single-user HRI the person and the robot face each other and hence a camera mounted onto the robot head provides high-resolution frontal images of the user's face such that head pose and eye orientation can both be easily and robustly estimated. In the case of multi-party HRI the eyes are barely detected since the participants often turn their faces away from the camera. Consequently, VFOA estimation methods based on eye detection and eye tracking are ineffective and one has to estimate the VFOA, indirectly, without explicit eye detection.

We propose a Bayesian switching dynamic model for the estimation and tracking gaze directions and VFOAs of several persons involved in social interaction. While it is assumed that head poses (location and orientation) and target locations can be directly detected from the data, the unknown gaze directions and VFOAs are treated as latent random variables. The proposed temporal graphical model, that incorporates gaze dynamics and VFOA transitions, yields (i) a tractable learning algorithm and (ii) an efficient gaze-and-VFOA tracking method.\footnote{Supplementary materials, that include a software package and examples of results, are available at \url{https://team.inria.fr/perception/research/eye-gaze/}.} The proposed method may well be viewed as a computational model of \cite{Freedman1997,freedman2008coordination}.

%\textcolor{red}{Next paragraph in section implementation ?}

The method is evaluated using two publicly available datasets, \emph{Vernissage}~\cite{Jayagopi2012} and \emph{LAEO} \cite{Marin-Jimenez2014}. These datasets consist of several hours of video containing situated dialog between two persons and a robot (\emph{Vernissage}) and human-human interactions (\emph{LAEO}). We are particularly interested in finding participants that either gaze to each other, gaze to the robot, or gaze to an object. \emph{Vernissage} is recorded with a motion capture system (a network of infrared cameras) and with a camera placed onto the robot head. \emph{LAEO} is collected from TV shows.

%Experimental results and a benchmark conducted with this dataset show that our method performs well and yields state-of-the-art results.

The remainder of this paper is organized as follows. Section~\ref{sec:related_works} provides an overview of related work in gaze, VFOA and head-pose estimation. Section~\ref{sec:model} introduces the paper's mathematical notations and definitions, states the problem formulation and describes the proposed model. Section~\ref{sec:inference} presents in detail the model inference and Section ~\ref{sec:learning} derives the learning algorithm. Section~\ref{sec:implementation} provides implementation details and Section~\ref{sec:results} describes the experiments and reports the results.

% The remainder of the paper is organized as follows. Section~\ref{sec:problem-formulation} formulates VFOA and gaze estimation as a MAP problem and describes the associated graphical model. Section~\ref{sec:observation-dynamics} describes the likelihood model and derives the gaze and the VFOA dynamics. Section~\ref{sec:inference} show how the MAP problem is cast into a switching Kalman filter formulation and describes the associated parameter learning method. Section~\ref{sec:experiments} describes in detail experiments conducted with the \emph{Vernissage} dataset. Finally, Section \ref{sec:conclusions} draws some conclusions.

%% file: section_related_works.tex
\section{Related Work}
\label{sec:related_works}
%\textcolor{red}{To be updated}

As already mentioned, the VFOA is correlated with gaze.
Several methods proceed in two steps, in which the gaze direction is estimated first, and then used to estimate VFOA. In scenarios that rely on precise estimation of gaze~\cite{Yu2004,Toyama2012} a head-mounted camera, like the one in~\cite{Hong2012}, can be used to detect the iris with high accuracy.
%Coupled with 3D eye shape extraction, the eyeball orientation can then be easily retrieved.
Head-mounted eye trackers provide extremely accurate gaze measurements and in some circumstances eye-tracking data can be used to estimate objects of interest in videos \cite{Kurzhals17visual}. Nevertheless, they are invasive instruments and hence not appropriate for analyzing social interactions.

Gaze estimation is relevant for a number of scenarios, such as  car driving~\cite{Smith2003} or interaction with smartphones~\cite{Krafka2016}. In these situations, either the field of view is limited, hence the range of gaze directions is constrained (car driving), or active human participation ensures that the device yields frontal
views of the user's face, thus providing accurate eye measurements \cite{Hong2012,Smith2003,Matsumoto2000,Ohno2004}. In some scenarios the user is even asked to limit head movements~\cite{Lu2014}, or to proceed through a calibration phase~\cite{Ohno2004,Lu2014b}. Even if no specific constraints are imposed, single-user scenarios inherently facilitate the task of eye measurement~\cite{Matsumoto2000}. At the best of our knowledge, there is no gaze estimation method that can deal with unconstrained scenarios, \eg participants not facing the cameras, partially or totally occluded eyes, etc. In general, eye analysis is inaccurate when participants are faraway from the camera.

An alternative is to approximate gaze direction with head pose \cite{Murphy2009}. Unlike eye-based methods, head pose can be estimated from low-resolution images, \ie distant cameras \cite{zabulis20093d,chamveha2013head,rajagopal2014exploring,yan2016multi,qin2016social}.
%Multiple-camera settings are not always available.
These methods estimate gaze only approximatively since eyeball orientation can differ from head orientation by $\pm 35\degree$~\cite{Stahl1999}. Gaze estimation from head orientation can benefit from the observation that gaze shifts are often achieved by synchronously moving the head and the eyes~\cite{goossens1997human,Freedman1997,freedman2008coordination}. The correlation between head pose and gaze has also been exploited in~\cite{Stiefelhagen2002a}.
  More recently,~\cite{Lanillos2017} combined head and eye features to estimate the gaze direction using an RGB-D camera. The method still requires that both eyes are visible.

Several methods were proposed to infer VFOAs either from gaze directions~\cite{Asteriadis2014}, or from head poses~\cite{Marin-Jimenez2014,Ba2009,Sheikhi2012,Yucel2013}. For example, in \cite{Marin-Jimenez2014} it is proposed to build a gaze cone around the head orientation and targets lying inside this cone are used to estimate the VFOA. While this method was successfully applied to movies, its limitation resides in its %poor accuracy, in particular when two targets are close to each other.
vagueness: the VFOA information is limited to whether there are two people looking at each other or not.

An interesting application of  VFOA estimation it the analysis of social behavior of participants engaged in meetings, \eg ~\cite{Stiefelhagen2002a,Ba2009,Otsuka2006,Duffner2015}. Meetings are characterized by interactions between seated people that interact based on speech and on head movements.  Some methods estimate the most likely VFOA associated with a head orientation \cite{Stiefelhagen2002a,Otsuka2006}. The drawback of these approaches is that they must be purposively trained for each particular meeting layout.
The correlation between VFOA and head pose was also investigated in \cite{Ba2009} where an HMM is proposed  to infer VFOAs from head and body orientations.  This work was extended to deal with more complex scenarios, such as participants interacting with a robot \cite{Sheikhi2012,Sheikhi2015}.
An input-output HMM is proposed in \cite{Sheikhi2015} to enable to model the following contextual information: participants tend to look to the speaker, to the robot, or to an object which is referred to by the speaker or by the robot. The results of \cite{Sheikhi2015} show that this improves the performance of VFOA estimation. Nevertheless, this method requires additional information, such as speaker identification or speech recognition.

% The pioneering work of~\cite{Ba2009} proposed a geometric formulation inspired from cognitive model to estimate gaze direction from head pose, independently of the situation. By introducing a reference direction, they assumed a linear mapping between head, gaze and reference direction. With the correct choice of reference direction, this model generates realistic gaze direction \wrt head direction, in accordance to psychophysics experiments. This model was extended in~\cite{Sheikhi2012} in which the reference direction could vary over time, and later in~\cite{Sheikhi2015} where additional contextual information are used.

The problem of joint estimation of gaze and of VFOA was addressed in a human-robot cooperation task~\cite{Yucel2013}. In such a scenario the user doesn't necessarily face the camera and robot-mounted cameras have low-resolution, hence the estimation of gaze from direct analysis of eye regions is not feasible. \cite{Yucel2013} proposes to learn a regression between  the space of head poses and the space of gaze directions and then to predict an unknown gaze from an observed head pose. The head pose itself is estimated by fitting a 3D elliptical cylinder to a detected face, while the associated gaze direction corresponds to the 3D line joining the head center to the target center. This implies that during the learning stage, the user is instructed to gaze at targets lying on a table in order to provide training data. The regression parameters thus estimated correspond to a discrete set of head-pose/gaze-direction pairs (one for each target); an erroneous gaze may be predicted when the latter is not in the range of gaze directions used for training.

%\textcolor{red}{Next paragraph in the introduction ?}

A summary of the  proposed Bayesian dynamic model and experiments with the \emph{Vernissage}~\cite{Jayagopi2012} motion capture dataset were presented in \cite{masse2016simultaneous}. In this article we provide a detailed and comprehensive description and analysis of the proposed model, of the model inference, of the learning methodology, and of the associated algorithms. We show results with both motion capture and RGB data from \emph{Vernissage}. Additionally, we show results with the \emph{LAEO} dataset \cite{Marin-Jimenez2014}.

%% file: section_model.tex
\section{Proposed Model}
\label{sec:model}

\begin{figure*}[t!]
  \begin{center}
  \begin{tabular}{cc}
    \includegraphics[width=0.45\linewidth]{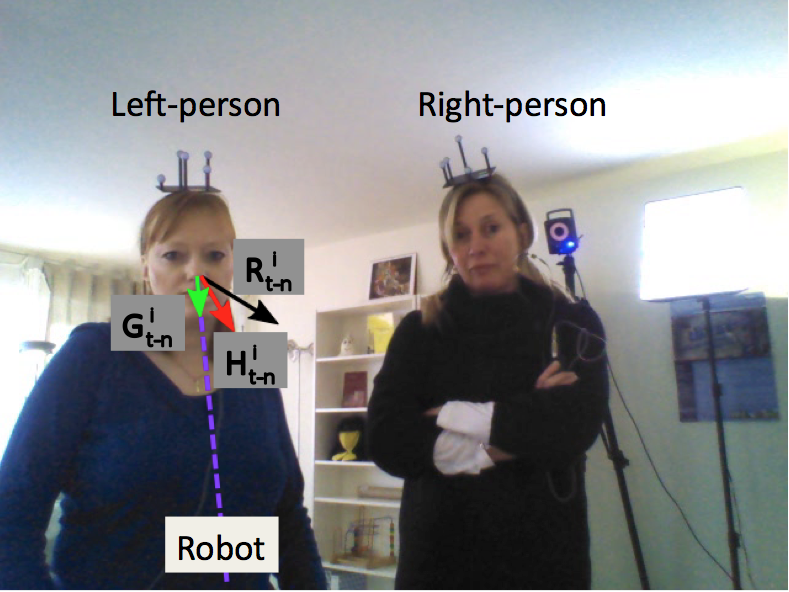}
   &
    \includegraphics[width=0.45\linewidth]{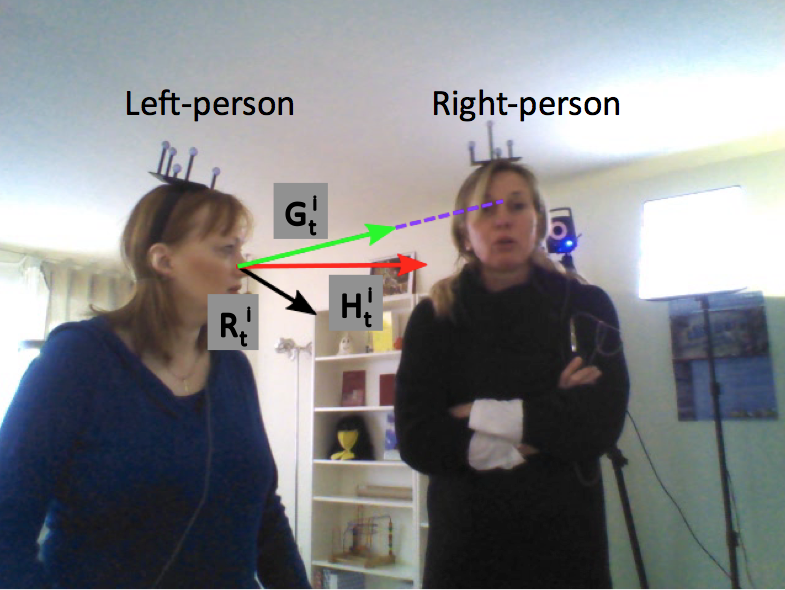}
    \end{tabular}
  \end{center}
  \caption{\label{fig:hgr} This figure illustrates the principle of our method and displays the observed and latent variables associated with a person (\textit{left-person} indexed by $i$). The two images were grabbed with a camera mounted onto the head of a robot and they correspond to frames $t-n$ (left image) and $t$ (right image), respectively. The following variables are displayed: head orientation (red arrow), $\H_{t-n}^i, \H_{t}^i$ (observed variables), as well as the latent variables estimated with the proposed method, namely gaze direction (green arrow), $\G_{t-n}^i,\G_{t}^i$,  VFOA, $\VV_{t-n}^i,\VV_{t}^i$, and head reference orientation (black arrow), $\R_{t-n}^i,\R_{t}^i$ (that coincides with upper-body orientation). In this example \textit{left-person} gazes towards the \textit{robot} at $t-n$, then turns her head to eventually gaze towards \textit{right-person} at $t$, hence her VFOA switches from $\VV_{t-n}^i=\textit{robot}$ to $\VV_{t}^i=\textit{right-person}$.
%The proposed model gathers the dynamics of the three latent variables (gaze direction, head reference orientation, and VFOA) into a switching Kalman filter formulation.
%namely the gaze direction $\G$ (latent), the head orientation $\H$ (observed) and the head reference orientation $\R$ (observed). In this example, the gaze of person $i$  is aligned with $\X_t^{ij}$ (dashed line), which is the direction from person $i$ to the person on the right (indexed $j$), hence the visual focus of attention of $i$ is equal to $j$, namely $\VV_t^i=j$.
}
\end{figure*}

The proposed mathematical model model is inspired from psychophysics~\cite{Freedman1997,freedman2008coordination}.
In unconstrained scenarios a person switches his/her gaze from one target to another target, possibly using both head and eye movements. Quick eye movements towards a desired object of interest are called saccades. Eye movements can also be caused by the vestibule-ocular reflex that compensates for head movements such that one can maintain his/her gaze in the direction of the target of interest. Therefore, in the general case, gazing to an object is achieved by a combination of eye and head movements.

In the case of small gaze shifts, \eg reading or watching TV, eye movements are predominant.
In the case of large gaze shifts, often needed in social scenarios, head movements are necessary since
eyeball movements have limited range, namely $\pm$35\textdegree{} \cite{Stahl1999}. Therefore, the proposed model considers that gaze shifts are produced by head movements that occur simultaneously with eye movements.

%~\textcolor{red}{The next paragraph has been added to sum up the technical parts and inform about the existence of the algorithm later (answers to questions from R3 and R4). Should it be in the introduction ?}
%
%  The rest of this section describes a mathematical embedding for representing VFOA, gaze and head pose of multiple people interacting. It proposes a model to take into account their dependencies. In this framework, a filtering procedure is derived in section~\ref{sec:inference}. This provides a method to sequentially process new input data (head poses) and to combine them with current knowledge to estimate the distribution of the output data (gaze directions and VFOAs). The corresponding algorithm is explicitly set out in section~\ref{subsec:implementation}. Finally, section~\ref{sec:learning} proposes a procedure to learn the unknown parameters from the model.

\subsection{Problem Formulation}
\label{subsec:problem_formulation}

We consider a scenario composed of $N$ active targets and $M$ passive targets. An active target is likely to move and/or to have a leading role in an interaction. Active targets are persons and robots.\footnote{Note that in case of a robot, the gaze direction and the head orientation are identical and that the latter can be easily estimated from the head motors.} Passive targets are objects, \eg  wall paintings. The set of all targets is indexed from $0$ to $N+M$, where the index $0$ designates ``no target". Let $i$ be an active target (a person or a robot), $1\leq i\leq N$, and $j$ be a passive target (an object), $N+1\leq j\leq N+M$. A VFOA is a discrete random variable defined as follows: $\VV_t^i = j$ means \textit{person (or robot) $i$ looks at target $j$ at time $t$}. The VFOA of a person (or robot) $i$ that looks at none of the known targets is $\VV_t^i=0$. The case $\VV_t^i=i$ is excluded. The set of all VFOAs at time $t$ is denoted by $\V_t = \left( \VV_t^1, \ldots, \VV_t^N \right)$.

Two continuous variables are now defined: head orientation and gaze direction. The head orientation of person $i$ at $t$ is denoted with $\H_t^i = [\phi_{H,t}^i, \theta_{H,t}^i]^\top$, \ie the pan and tilt angles of the head with respect to some fixed coordinate frame. %The roll angle can be estimated as well but it is not considered here because the roll's axis is the gaze direction itself.
The gaze direction of person $i$ is denoted with $\G_t^i$ and is also parameterized by pan and tilt with respect to the same coordinate frame, namely $\G_t^i = [\phi_{G,t}^i, \theta_{G,t}^i]^\top$. Although eyeball orientation is neither needed nor used, it is worth noticing that it is the difference between $\G_t^i$ and $\H_t^i$. These variables are illustrated on Fig.~\ref{fig:hgr}.

Finally, to establish a link between VFOAs and  gaze directions, the target locations must be defined as well. Let $\X_t^i=[x_t^i, y_t^i, z_t^i]^\top$ be
the location of target $i$. In the case of a person, this location corresponds to the head center while in the case of a passive target, it corresponds to the target center. These locations are defined in the same coordinate frame as above. Also notice that the direction from the active target $i$ to target $j$ is defined by the unit vector $\X_t^{ij}= (\X_t^j - \X_t^i)/ \| \X_t^j - \X_t^i \|$ which can also be parameterized by two angles, $\X_t^{ij}= [\phi_{X,t}^{i,j}, \theta_{X,t}^{i,j}]^\top$.

As already mentioned, target locations and head orientations are observed random variables, while VFOAs and gaze directions are latent random variables. The problem to be solved can now be formulated as a maximum a posteriori (MAP) problem:
\begin{align}
  \hat{\V}_t, \hat{\G}_t &= \argmax_{\V_t,\G_t} P(\V_t,\G_t | \H_{1:t}, \X_{1:t})\label{eq:goal}
\end{align}

%% \begin{figure}[t]
%%   \begin{center}
%%     \includegraphics[width=0.95\linewidth]{0105_hgr_nao}
%%     % \includegraphics[width=0.55\linewidth]{0105_hgr_nao}
%%     % \includegraphics[width=0.40\linewidth]{0105_hgr_top}
%%   \end{center}
%%   \caption{\label{fig:hgr}
%%   This figure illustrates the main variables associated with a person, \eg the person on the left (indexed $i$), namely the gaze direction $\G$ (latent), the head orientation $\H$ (observed) and the head reference orientation $\R$ (observed). In this example, the gaze of person $i$  is aligned with $\X_t^{ij}$ (dashed line), which is the direction from person $i$ to the person on the right (indexed $j$), hence the visual focus of attention of $i$ is equal to $j$, namely $\VV_t^i=j$.}
%% \end{figure}

Since there is no deterministic relationship between head orientation and gaze direction, we propose to model it  probabilistically. For this purpose, we introduce an additional latent random variable, namely the head \textit{reference} orientation, $\R_t^i = [\phi_{R,t}^i, \theta_{R,t}^i]^\top$, which we choose to coincide with the upper-body orientation. We use the following generative model, initially introduced in~\cite{Ba2009}, linking gaze direction, head orientation, and head reference orientation:
\begin{align}
\label{eq:expectation_H}
  P(\H_t^i | \G_t^i, \R_t^i; \alphabf, \Sigmabf_{\H}) & =  \mathcal{N} (\H_t^i ; \mubf_{\H,t}^i, \Sigmabf_{\H}), \\
  \label{eq:mean_H}
   \textrm{with} \quad \mubf_{\H,t}^i & = \alphabf\G_t^i + (\I_2 - \alphabf)\R_t^i,
\end{align}
where $\Sigmabf_{\H}\in\mathbb{R}^{2\times 2}$ is a covariance matrix, $\I_2\in\mathbb{R}^{2\times 2}$ is the identity matrix and $\alphabf=\diag ( \alpha_1, \alpha_2)$ is a diagonal matrix of mixing coefficients, $0 < \alpha_1,\alpha_2 <1$. Also it is assumed that the covariance matrix is the same for all the persons and over time.
Therefore, head orientation is an observed random variable normally distributed around a convex combination between two latent variables: gaze direction and head reference orientation.

%In a Bayesian dynamic model, one has to define the state dynamics. In our case, the state is defined by a discrete variable and by two continuous variables. We start by specifying the dynamics of the two continuous variables.

\subsection{Gaze Dynamics}

The following model is proposed:
\begin{align}
\label{eq:expectation_G}
P(\G_t^i | \G_{t-1}^i\Gdot_{t-1}^i,\VV_t^i=j,\X_t) & =  \mathcal{N} (\G_t^i ; \mubf_{\G,t}^{ij} , \Gammabf_{\G}),\\
\label{eq:distribution_Gdot}
P(\Gdot_t^i |  \Gdot_{t-1}^i) &=    \mathcal{N}(\Gdot_t^i;\Gdot_{t-1}^i,\Gammabf_\Gdot),
\end{align}
with:
\begin{align}
\label{eq:mean_G}
\mubf_{\G,t}^{ij}  =
\begin{cases}
 \G_{t-1}^i + \Gdot_{t-1}^i \: dt, & \mbox{if} \: j = 0,\\
 \betabf\G_{t-1}^i + (\I_2-\betabf) \X_t^{ij} + \Gdot_{t-1}^i \: dt, & \mbox{if} \: j \neq 0,
\end{cases}
\end{align}
where $\Gdot_t^i = d\G_t^i/dt$ is the gaze velocity,
$\Gammabf_{\G},\Gammabf_\Gdot \in\mathbb{R}^{2\times 2}$ are covariance matrices, and
$\betabf=\diag ( \beta_1, \beta_2)$ is a diagonal matrix of mixing coefficients, $0 < \beta_1,\beta_2 <1$.
% Radu a enlevŽ ce texte car il est faux!
%This formulation can be interpreted as follows. First, if $\VV_t^i \neq 0$, the gaze direction of person $i$ at $t$ is normally distributed around a convex combination between the gaze direction of $i$ at $t-1$ and the direction from person $i$ to target $j$. Second, if $\VV_t^i = 0$, the gaze direction of person $i$ at $t$ is normally distributed around his/her gaze direction at $t-1$.
Therefore, if a person looks at one of the  targets, then his/her gaze dynamics depends on the person-to-target direction $\X_t^{ij}$ at a rate equal to $\betabf$, and if a person doesn't look at one of the targets, then his/her gaze dynamics follows a random walk.

\begin{figure}[t]
\centering
\input{tikz_complete_model}
\caption{
Graphical representation showing the dependencies between the variables of the proposed Bayesian dynamic model. The discrete latent variables (visual focus of attention) are shown with squares while continuous variables are shown with circles: observed variables (head pose and target locations) are shown with shaded circles and latent variables (gaze and reference) are shown with white circles.
}
\label{fig:complete_model}
\end{figure}
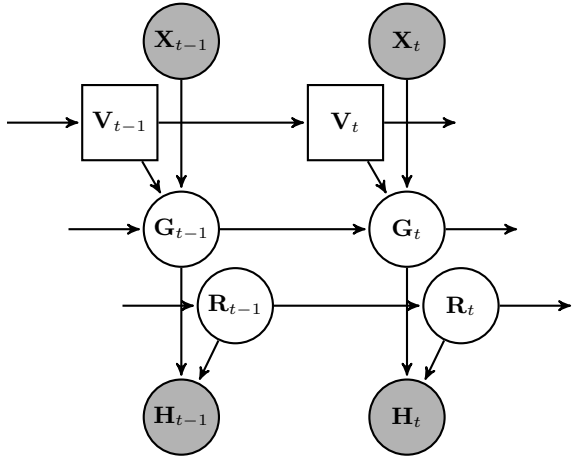

The head reference orientation dynamics can be defined in a similar way:
\begin{align}
\label{eq:expectation_R}
P(\R_t^i | \R_{t-1}^i,\Rdot_{t-1}^i) &= \mathcal{N} (\R_t^i ; \mubf_{\R,t}^i, \Gammabf_{\R}), \\
\label{eq:distribution_Rdot}
P(\Rdot_t^i |  \Rdot_{t-1}^i) &=    \mathcal{N}(\Rdot_t^i;\Rdot_{t-1}^i,\Gammabf_\Rdot), \\
\textrm{with} \; \mubf_{\R,t}^i &= \R_{t-1}^i + \Rdot_{t-1}^i \: dt, \nonumber
\end{align}
where $\Rdot_t^i = d \R_t^i/dt$ is the head reference orientation velocity and
$\Gammabf_{\R},\Gammabf_\Rdot \in\mathbb{R}^{2\times 2}$ are covariance matrices.
The dependencies between all the model variables are shown as a graphical representation in Figure~\ref{fig:complete_model}.

%% FIGURE

%%%% PART FUSED FROM PREVIOUS SECTION 'PROPOSED MODEL'

%\section{The Proposed Model}

%So far the problem to be solved was formulated and the model variables were defined. Considering the above relationships as probabilistic expectations, one is left with a Bayesian network, and the objective function \eqref{eq:goal} can be solved as a maximum likelihood problem.
% More specifically, equations~\ref{eq:expectation_H},~\ref{eq:expectation_R} and~\ref{eq:expectation_G} become
%In particular, we consider $\H$, $\G$ and $\R$ to be the individual dynamics.
  %% Given VFOAs along time (see below), we assume that gaze directions, head orientations and head reference orientations are independent.
%  Additionally, the use of a high-level VFOA variable is convenient to model the interactions between people. The decision to look a particular target ($\VV_{t+1}^i$) depends on the VFOA of other people ($\V_t$). Yet, once all the VFOAs are known, we assume that gaze directions from different people, $\G_t^i$ and $\G_t^j$, are independent given $\V_{1:t}$. Similarly, head orientations and head reference orientations from different people are independent given $\V_{1:t}$. %% For instance, if $i\neq j$, $\G_t^i \perp\!\!\!\perp \G_t^j\ | \V_{1:t}$, and similarly for $\H_t^i$ and $\R_t^i$.
It is assumed that the gaze directions associated with different people are independent, given the VFOAs  $\V_{1:t}$. The cross-dependency between different people is taken into account by the VFOA dynamics as detailed in section~\ref{subsec:vfoa_dynamics} below. Similarly, head orientations, and head reference orientations associated with different people are independent, given the VFOAs.
By combining the above equations with this independency assumption, we obtain:
\begin{align}
  P(\H_t | \G_t, \R_t) &= \prod_i \mathcal{N}(\H_t^i; \mubf_{\H,t}^i , \Sigmabf_\H)\label{eq:distribution_H}\\
    P(\G_t | \G_{t-1},\Gdot_{t-1}, \V_t, \X_t) &= \prod_{i,j} \mathcal{N}(\G_t^i; \mubf_{\G,t}^{ij} , \Gammabf_\G)^{\delta_j(\V_t^i)}\label{eq:distribution_G}\\
    P(\R_t | \R_{t-1},\Rdot_{t-1}) &= \prod_i \mathcal{N}(\R_t^i; \mubf_{\R,t}^i , \Gammabf_\R)\label{eq:distribution_R}
  % P(\V_t   | \V_{t-1}) &= f(\V_{t-1})\label{eq:distribution_V}
\end{align}
where the dependencies between variables are embedded in the variable means, \ie \eqref{eq:mean_H} and \eqref{eq:mean_G}.
%The notation $\delta_j(\VV_t^i)$ corresponds to a Kronecker delta: $\delta_j(\VV_t^i)=1$ when $\VV_t^i=j$, and $\delta_j(\VV_t^i)=0$ otherwise. It is used for dynamics selection.
The covariance matrices will be estimated via training. While gaze directions can vary a lot, we assume that head reference orientations are almost constant over time, which can be enforced by imposing that the total variance of gaze is much larger than the total variance of head reference orientation, namely:

%In other terms, The head reference orientation is supposed to be the orientation around which the gaze varies, hence the former should seldom change.
%We propose to enforce this relative dependency with the following constraint:
\begin{equation}
\label{eq:covariance-change}
\Tr (\Gammabf_\G) \gg \Tr (\Gammabf_\R),
\end{equation}
%  The trace of a covariance matrix is used here as an approximation of the variance for multidimensional variables. As the reference orientation is more stable than gaze direction, its variance is lower.

%\begin{figure*}[t]
%  \centering
%  \includegraphics[width=\linewidth]{vfoa_pipeline_passive}
%  \caption{Pipeline of the complete system using RGB Data. First, the inputs $\H_t$ and $\X_t$ are extracted from the image when possible, then they are combined with current state representation through an update (filtering) step described in section~\ref{sec:inference}. Section~\ref{sec:learning} proposes a method to learn the parameters $\mathbf{p}$, $\alphabf$, $\betabf$, $\Gammabf_{\LL}$, $\Sigmabf_{\H}$.}
%  \label{fig:pipeline}
%\end{figure*}
%

\subsection{VFOA Dynamics}
\label{subsec:vfoa_dynamics}

Using a first-order Markov approximation, the VFOA transition probabilities can be written as:
\begin{equation}
\label{eq:vfoa-definition}
P(\V_t | \V_{1:t-1}) = P(\V_t | \V_{t-1}),
\end{equation}
Notice that matrix $P(\V_t | \V_{t-1})$ is of size $(N+M)^N \times (N+M)^N$. Indeed, there are $N$  persons (active targets), and $N+M+1$ targets (one "no" target, N active targets and $M$ passive targets) and the case of a person that looks to him/herself is excluded. For example, if $N=2$ and $M=4$, matrix \eqref{eq:vfoa-definition} has $(2+4)^{2\times2} = 1296$ entries.
The estimation of this matrix  would require, in principle, a large amount of training data, in particular in the presence of many symmetries. We show that, in practice, only 15 different transitions are possible. This can be seen on the following grounds.

We start by assuming conditional independence between the VFOAs at $t$:
\begin{align}
  P(\V_t|\V_{t-1}) &= \prod_i P(\VV_t^i | \V_{t-1}).\label{eq:independence_V}
\end{align}
Let's consider $V_t^i$, the VFOA of person $i$ at $t$, given $\V_{t-1}$, the VFOAs at $t-1$. One can distinguish two cases:
\begin{itemize}
\item $V_{t-1}^i=k$ where $k$ is either a passive target, $N<k\leq N+M$, or it is none of the targets, $k=0$;  in this case $V_{t}^i$ depends only on $V_{t-1}^i$, and
\item $V_{t-1}^i=k$, where $k\neq i$ is a person $1\leq k \leq N$; in this case $V_{t}^i$ depends on the  both $V_{t-1}^i$ and $V_{t-1}^k$.
\end{itemize}
To summarize, we can write that:
\begin{align}
\label{eq:vfoa-transition-person-i}
 P(& \VV_t^i =j | \V_{t-1})  = \nonumber \\
&
 \begin{cases}
   P(\VV_t^i =j| \VV_{t-1}^i=k,\VV_{t-1}^k=l) & \text{if} \; 1 \leq k \leq N, \\
 P(\VV_t^i =j| \VV_{t-1}^i=k) & \text{otherwise}.
 \end{cases}
\end{align}

Based on this it is now possible to count the total number of possible VFOA transitions. With the same notations as in \eqref{eq:vfoa-transition-person-i}, we have the following possibilities:
\begin{itemize}
\item $k=0$ (no target): there are two possible transitions, $j=0$ and $j\neq 0$.
\item $N < k \leq N+M$ (passive target): there are three possible transitions, $j=0$, $j=k$, and $j \neq k$.
\item $1\leq k \leq N, l=0$ (active target $k$ looks at no target): there are three possible transitions, $j=0$, $j=k$, and $j \neq k$.
\item $1\leq k \leq N, l=i$ (active target $k$ looks at person $i$): there are three possible transitions, $j=0$, $j=k$, and $j \neq k$.
\item $1\leq k \leq N, l\neq 0,i$ (active target $k$ looks at active target $l$ different than $i$): there are four possible transitions, $j=0$, $j=k$, $j=l$ and $j \neq k,l$.
\end{itemize}
Therefore, there are 15 different possibilities for $P(\VV_t^i =j | \V_{t-1})$, \ie
%appendix~A.
appendix~\ref{app:transitions_list}.
Moreover, by assuming that the VFOA transitions don't depend on $i$, we conclude that the transition matrix may have up to 15 different entries. Moreover, the number of possible transitions is even smaller if there is no passive target ($M=0$), or if the number of active targets is small, \eg $N<3$.
This considerably simplifies the task of estimating this matrix and makes the task of learning tractable.

%% file: tikz_complete_model.tex
\begin{tikzpicture}[->,>=stealth',shorten >=1pt,auto,node distance=2.5cm,inner sep=0pt, minimum size=1cm, thick,main node/.style={circle,draw,font=\sffamily\small\bfseries}]

  \node[main node] (gt) {$\G_{t-1}$};
  \node[main node] (gt1) [right of=gt, xshift=0.5cm] {$\G_t$};
  \node[main node, fill=black!30!white] (ht) [below of=gt] {$\H_{t-1}$};
  \node[main node, fill=black!30!white] (ht1) [below of=gt1] {$\H_t$};
  \node[main node] (rt) [below right of=gt, xshift=-1.05cm, yshift=0.75cm] {$\R_{t-1}$};
  \node[main node] (rt1) [below right of=gt1, xshift=-1.05cm, yshift=0.75cm] {$\R_t$};
  \coordinate[left of=gt, xshift=1cm] (gtm1);
  \coordinate[left of=rt, xshift=1cm] (rtm1);
  \coordinate[right of=gt1, xshift=-1cm] (gt2);
  \coordinate[right of=rt1, xshift=-1cm] (rt2);
  \node[main node, fill=black!30!white] (xt) [above of=gt] {$\X_{t-1}$};
  \node[main node, fill=black!30!white] (xt1) [above of=gt1] {$\X_t$};
  \node[main node, rectangle] (vt) [above left of=gt, xshift=0.95cm, yshift=-0.35cm] {$\V_{t-1}$};
  \node[main node, rectangle] (vt1) [above left of=gt1, xshift=0.95cm, yshift=-0.35cm] {$\V_t$};
  \coordinate[left of=vt, xshift=1cm] (vtm1);
  \coordinate[right of=vt1, xshift=-1cm] (vt2);

%  \node[main node, fill=black!30!white, minimum size=0.5cm] (greycircle) [below left of=ht, yshift=0.9cm] {};
%  \node[main node, minimum size=0.5cm] (whitecircle) [below of=greycircle, yshift=1.9cm] {};
%  \node[main node, rectangle, minimum size=0.5cm] (whiterectangle) [below of=whitecircle, yshift=1.9cm] {};

%  \node[right = of greycircle,     xshift=-2.3cm] {Continuous observed variables};
%  \node[right = of whitecircle,    xshift=-2.3cm] {Continuous latent variables};
%  \node[right = of whiterectangle, xshift=-2.3cm] {Discrete latent variables};

  \path[every node/.style={font=\sffamily\small}]
      (gt)  edge node {} (ht)
            edge node {} (gt1)
      (rt)  edge node {} (ht)
            edge node {} (rt1)
      (rt1) edge node {} (ht1)
            edge node {} (rt2)
      (gt1) edge node {} (ht1)
            edge node {} (gt2)
      (gtm1)edge node {} (gt)
      (rtm1)edge node {} (rt)
      (xt)  edge node {} (gt)
      (xt1) edge node {} (gt1)
      (vt)  edge node {} (gt)
            edge node {} (vt1)
      (vt1) edge node {} (gt1)
            edge node {} (vt2)
      (vtm1)edge node {} (vt)
      ;
\end{tikzpicture}

%% file: section_inference.tex
\section{Inference}
\label{sec:inference}

We start by simplifying the notation, namely  $\LL_t = [\G_t;\dot{\G}_t;\R_t;\dot{\R}_t]$
where $[\cdot;\cdot]$ denotes vertical concatenation.
The emission probabilities~\eqref{eq:distribution_H} become:
\begin{align}
\label{eq:distribution_HL}
  P(\H_t | \LL_t) &= \prod_i \mathcal{N}(\H_t^i; \mubf_{\H,t}^i, \Sigmabf_\H),\\
  \label{eq:expectation_HL}
  \textrm{with}\quad \mubf_{\H,t}^i &= \C \LL_t^i,
  \end{align}
where %$2\times 8$
matrix $\C$ is obtained from the definition of $\LL_t$ above and from~\eqref{eq:mean_H}:
  \begin{equation}
   \C=\begin{pmatrix}
   \alpha_1 & 0 & 0 & 0 & 1-\alpha_1 & 0 & 0 & 0 \\
  0 &  \alpha_2 & 0 & 0 & 0 & 1-\alpha_2 & 0 & 0
    \end{pmatrix}.\nonumber
\end{equation}

The transition probabilities can be obtained by combining~\eqref{eq:distribution_G} and~\eqref{eq:distribution_R} with~\eqref{eq:distribution_Gdot} and~\eqref{eq:distribution_Rdot}:
\begin{align}
\label{eq:distribution_L}
  P(\LL_t | \V_t, \LL_{t-1}, \X_t) &= \prod_i\prod_j \mathcal{N}(\LL_t^i; \mubf_{\LL,t}^{ij}, \Gamma_\LL)^{\delta_j(\VV_t^i)},\\
   \label{eq:expectation_L}
  \textrm{with}\quad \mubf_{\LL,t}^{ij} &= \A_t^{ij} \LL_{t-1}^i + \bb_t^{ij}\\
  \label{eq:covariance_L}
  \textrm{and}\quad \Gammabf_{\LL} &= \begin{pmatrix}\Gammabf_{\G}&&&\\&\Gammabf_{\Gdot}&&\\&&\Gammabf_{\R}&\\&&&\Gammabf_{\Rdot}\end{pmatrix},
\end{align}
where $\A_t^{ij}$ is an $8\times 8$ matrix and $\bb_t^{ij}$ is an $8\times 1$ vector. The indices $i$,$j$ and $t$ cannot be dropped since the transitions depend on $\X_t^{ij}$ from~\eqref{eq:mean_G}.

The MAP problem~\eqref{eq:goal} can now be derived in a Bayesian framework for the VFOA variables:
\begin{align}
  P(\V_t|\H_{1:t},\X_{1:t}) &= \int  P(\V_t,\LL_t|\H_{1:t},\X_{1:t}) d\LL_t.
\end{align}
We propose to study the filtering distribution of the joint latent variables, namely $ P(\V_t,\LL_t|\H_{1:t},\X_{1:t})$. Indeed, Bayes rule yields:
\begin{align}
  P(\V_t,\LL_t|\H_{1:t},\X_{1:t})&= \frac{1}{c} P(\H_t | \LL_t) P(\LL_t,\V_t|\H_{1:t-1}, \X_{1:t}). \label{eq:joint_latent_bayes}
\end{align}
where $c$ is the normalization evidence. Now we can introduce $\V_{t-1}$ and $\LL_{t-1}$ using the sum rule:
\begin{align}
  P(\LL_t,& \V_t|      \H_{1:t-1}, \X_{1:t}) \nonumber\\
  & =\sum_{\V_{t-1}} \int P(\LL_t,\V_t,\LL_{t-1}, \V_{t-1}|\H_{1:t-1}, \X_{1:t}) d\LL_{t-1}\nonumber\\
  & =\sum_{\V_{t-1}} \int P(\LL_t|\V_t,\LL_{t-1},\X_t) P(\V_t|\V_{t-1})  \nonumber\\
  & \times P(\LL_{t-1}, \V_{t-1}|\H_{1:t-1}, \X_{1:t-1}) d\LL_{t-1}, \label{eq:joint_latent_sum_rule}
\end{align}
where unnecessary dependencies were removed. Combining~\eqref{eq:joint_latent_bayes} and~\eqref{eq:joint_latent_sum_rule} we obtain a recursive formulation in $P(\V_t,\LL_t|\H_{1:t},\X_{1:t})$. However, this model is still intractable without further assumptions. The main approximation used in this work consists of assuming local independence for the posteriors:
\begin{align}
  P(\LL_t, \V_t | \H_{1:t}, \X_{1:t}) &\simeq \prod_i P(\LL_t^i, \VV_t^i | \H_{1:t}, \X_{1:t}).\label{eq:independence_hypothesis}
\end{align}
\subsection{Switching Kalman Filter Approximation}
\label{subsec:SKF-approx}
Several strategies are possible, depending upon the structure of $P(\LL_t, \V_t | \H_{1:t}, \X_{1:t})$. Commonly used strategies to evaluate this distribution include variational Bayes or Monte-Carlo. Alternatively, we propose to cast the problem into the framework of switching Kalman filters (SKF)~\cite{Murphy1998}. We assume the filtering distribution to be Gaussian,
%% We propose a solution into the switching Kalman filter framework, that match locally the filtering distribution as a normal one
\begin{align}
  P(\LL_t, \V_t | \H_{1:t}, \X_{1:t}) &\propto \mathcal{N}(\LL_t; \mubf_t, \Sigmabf_t).\label{eq:skf_hypothesis}
\end{align}
From~\eqref{eq:independence_hypothesis} and~\eqref{eq:skf_hypothesis}  we obtain the following factorization:
\begin{equation}
  \label{eq:factorization-joint}
  P(\LL_t, \V_t | \H_{1:t}, \X_{1:t}) \propto \prod_i \prod_j \mathcal{N}(\LL_t^i; \mubf_t^{ij}, \Sigmabf_t^{ij})^{\delta_j(\VV_t^i)}.
\end{equation}
Thus,~\eqref{eq:joint_latent_sum_rule} can be split into $N$ components, one for each active target $i$:
\begin{align}
  P(\LL_t^i, &\VV_t^i=j | \H_{1:t},\X_{1:t}) \propto  P(\H_t^i|\LL_t^i)  \nonumber\\
   &\times \sum_{\V_{t-1}} \int \mathcal{N}(\LL_t^i; A_t^{ij}\LL_{t-1}^i + \bb_t^{ij}) P(\VV_t^i|\V_{t-1})\nonumber \\
  &\times \prod_k \mathcal{N}(\LL_{t-1}^i; \mubf_{t-1}^{ik}, \Sigmabf_{t-1}^{ik})^{\delta_k(\VV_{t-1}^i)} d\LL_{t-1}^i,
\end{align}
or, after several algebraic manipulations:
\begin{align}
  P(\LL_t^i, \VV_t^i=j | \H_{1:t},\X_{1:t}) \propto \sum_k w_{t-1,t}^{ijk} \mathcal{N}(\LL_t^i;\mubf_t^{ijk}, \Sigmabf_t^{ijk}).\label{eq:filtering_mixture}%\\
  % \mubf_t^{ijk} = \A_t^{ij}\mubf_{t-1}^{ik} + \bb_t^{ij} + \KK_t (\H_t^i - \A_t^{ij}\mubf_{t-1}^{ik} - \bb_t^{ij})
\end{align}
In this expression, $\mubf_t^{ijk}$ and $\Sigmabf_t^{ijk}$ are obtained by performing constrained Kalman filtering on $\mubf_{t-1}^{ik}$, $\Sigmabf_{t-1}^{ik}$ with transition dynamics defined by $\A_t^{ij}$ and $\bb_t^{ij}$, emission dynamics defined by $\C$, and observation $\H_t^i$, \ie \cite{Simon2010}. The weights $w_{t-1,t}^{ijk}$ are defined as $P(\VV_{t-1}^i=k \mid \VV_t^i=j, \H_{1:t},\X_{1:t})$. The constraint comes from the fact that $||\G_t^i-\H_t^i|| < 35\degree$ and is achieved by projecting the mean (refer to~\cite{Simon2010} for more details).

This can be rephrased as follows: from the filtering distribution at time $t-1$, there are $N+M$ possible dynamics for $\LL_t^i$. The normal distribution at time $t-1$ then becomes a mixture of $N+M$ normal distributions at time $t$ as shown in \eqref{eq:filtering_mixture}. % The mixture weights $c_{t-1,t}^{ijk}$ are defined as $P(\VV_t^i=j, \VV_{t-1}^i=k \mid \H_{1:t}\backslash\H_t^i,\X_{1:t})$.
However, we expect a single Gaussian such as $P(\LL_t^i, \VV_t^i=j | \H_{1:t},\X_{1:t}) \propto\mathcal{N}(\LL_t^i; \mubf_t^{ij}, \Sigmabf_t^{ij})$. This can be done by moment matching:
\begin{align}
  \mubf_t^{ij} &= \sum_k w_{t-1,t}^{ijk} \mubf_t^{ijk}\label{eq:update_mutij}\\
  \Sigmabf_t^{ij} &= \sum_k w_{t-1,t}^{ijk} (\Sigmabf_t^{ijk} + (\mubf_t^{ijk}-\mubf_t^{ij})(\mubf_t^{ijk}-\mubf_t^{ij})^\top)\label{eq:update_sigmatij}
\end{align}

Finally, it is necessary to evaluate $w_{t-1,t}^{ijk}$. Let's introduce the following notations:
\begin{align}
 c_{t-1,t}^{ijk} &= P(\VV_t^i=j, \VV_{t-1}^i=k|\H_{1:t},\X_{1:t}),\\
 c_t^{ij}      &= P(\VV_t^i=j |\H_{1:t},\X_{1:t}).
 \end{align}
It follows that
\begin{align*}
 c_t^{ij} = \sum_k c_{t-1,t}^{ijk}\quad \textrm{and}\quad w_{t-1,t}^{ijk} =\frac{c_{t-1,t}^{ijk}}{c_t^{ij}}.
\end{align*}
By applying Bayes formula to $c_{t-1,t}^{ijk}$, yields:
\begin{align}
  \label{eq:cijk}
  c_{t-1,t}^{ijk} \propto&\ P(\H_t| \VV_t^i=j, \VV_{t-1}^i=k,\H_{1:t-1},\X_{1:t})\nonumber\\
  %&\times %P(\VV_t^i=j, \VV_{t-1}^i=k, \H_{1:t-1},\X_{1:t-1})\\
  % \propto & f(\H_t) c_{t-1}^{ik} \sum_l c_{t-1}^{kl} P(\VV_t^i=j|\VV_{t-1}^i=k, \VV_{t-1}^k=l)
  \times&c_{t-1}^{ik} P(\VV_t^i=j|\VV_{t-1}^i=k, \H_{1:t-1}, \X_{1:t-1}) %\sum_l c_{t-1}^{kl} P(\VV_t^i=j|\VV_{t-1}^i=k,\VV_{t-1}^k=l).
\end{align}
Then, $c_{t-1}^{ik}$ is obtained from $c_{t-2,t-1}^{ijk}$ calculated at previous time step.
% by integrating out $\LL_{t-1}^i$ from $P(\VV_{t-1}^i, \LL_{t-1}^i|\H_{1:t-1},\X_{1:t-1})$, known from the previous time step.
The last factor in~\eqref{eq:cijk} is either equal to $\sum_l c_{t-1}^{kl} P(\VV_t^i=j|\VV_{t-1}^i=k, \VV_{t-1}^k=l)$ if $k$ is an active target, or $P(\VV_t^i=j|\VV_{t-1}^i=k)$ otherwise. Both cases are straightforward to compute. Finally, the first factor in~\eqref{eq:cijk}, the observation component, can be factorized as  $P(\H_t^i| \VV_t^i=j, \VV_{t-1}^i=k, \H_{1:t-1}, \X_{1:t}) \times \prod_{n\neq i}\sum_m \sum_p P(\H_t^n| \VV_t^n=m, \VV_{t-1}^n=p, \H_{1:t-1}, \X_{1:t})$. By introducing the latent variable $\LL$, we obtain:
\begin{align}
P& (\H_t^n|\VV_t^n=m,\VV_{t-1}^n=p,\H_{1:t-1}, \X_{1:t})\nonumber\\
&= \int P(\H_t^n|\LL_t^n)\ P(\LL_t^n|\LL_{t-1}^n, \VV_t^n=m, \X_t)\nonumber\\
      &\times P(\LL_{t-1}^n|\VV_{t-1}^n=p, \H_{1:t-1}, \X_{1:t-1}) d\LL_{t-1}^n d\LL_t^n.
\label{eq:introducing_l}
\end{align}
All the factors \eqref{eq:introducing_l}  are normal distributions, hence it integrates in closed-form.
%
% $P(\H_t^m| \H_{1:t-1},\X_{1:t})$ can also be solved in closed form by introducing $\VV_t^m$, $\VV_{t-1}^m$, $\LL_t^m$, $\LL_{t-1}^m$ through the sum rule.
In summary, we devised a procedure to estimate an online approximation of the joint filtering distribution of the VFOAs, $\V_t$, and of the gaze and head reference directions, $\LL_t$.

% Now, this distribution can be modelled as a mixture of normal distributions. The choice of which Gaussian to use depends on the value of $\V_t$: $P(\LL_t,\V_t)=P(\V_t)P(\LL_t|\V_t)$

% Let's assume further that $ P(\LL_{t-1}, \V_{t-1}|\H_{1:t-1}, \X_{1:t-1}) $ is the product of $N$ Gaussian Mixture Model (GMM), each having $N+M$ components. This is justified by the relation between the VFOA and the gaze direction. Namely, for one particular person, looking at one particular target, we want a unique prediction for gaze direction (up to gaussian noise uncertainty).

% which allows us to make use of the relationship between head direction, gaze direction, and head reference direction defined previously. Using variable independency assumptions, \ie Fig.~\ref{fig:complete_model}, the joint filtering distribution can be expanded as:
% \begin{align}
%   &P(\V_t, \G_t, \R_t | \H_{1:t},\X_{1:t}) \nonumber\\
%   &= \frac{P(\H_t | \G_t, \R_t) P(\V_t, \G_t, \R_t | \H_{1:t-1},\X_{1:t})}
%   {P(\H_t | \H_{1:t-1},\X_{1:t})}\label{eq:filtering_bayes}
% \end{align}
% which is composed of three terms: the observation likelihood $P(\H_t | \G_t, \R_t)$, the state predictive distribution $P(\V_t, \G_t, \R_t | \H_{1:t-1},\X_{1:t})$, and the observation predictive distribution $P(\H_t | \H_{1:t-1},\X_{1:t})$. The temporal dependency can also be introduced by the sum rule
% \begin{align}
%   &P(\V_t, \G_t, \R_t | \H_{1:t-1},\X_{1:t})=\\\nonumber
%   &\sum_{\V_{t-1}} \int P(\V_t, \G_t, \R_t | \V_{t-1}, \G_{t-1}, \R_{t-1}, \H_{1:t-1},\X_{1:t}) d\G_{t-1} d\R_{t-1}
% \end{align}

%% file: section_learning.tex
\section{Learning}
\label{sec:learning}

The proposed model has two sets of parameters that must be estimated: the transition probabilities associated with the discrete VFOA variables, and the parameters associated with the Gaussian distributions. Learning is carried out using $Q$ recordings with annotated VFOAs.
Each recording is composed of $T_q$ frames, $1\leq q\leq Q$ and contains $N_q$ active targets (the robot is the active target 1 and the persons are indexed from 2 to $N_q$) and $M_q$ passive targets.
  %%%% Submitted version
  %Both the head poses and the VFOAs of the active targets are available with all the frames.
  %%%% Proposition Benoit
  %  As previously, the head orientations $\H$ of active targets and the location $\X$ of both active and passive targets are available. Moreover, as opposed to the inference in which the goal is to estimate the VFOAs, it is supposed in this section that the VFOAs of the active targets are available to evaluating the model parameters.
  %%%% Proposition Radu
  %% The VFOAs are annotated in each frame, \eg the VFOA of active target $i$ is target $j$.
  %%%% Proposition Benoit
  In addition to target locations and head poses, it is worth noticing that the learning algorithm requires VFOA ground-truth annotations, while gaze directions are still treated as latent variables.

\subsection{Learning the VFOA Transition Probabilities}
\label{subsec:learning-vfoa-trans}

The VFOA transitions are drawn from the generalized Bernoulli distribution. Therefore, the transition probabilities can be estimated with $P(\VV_t^i=j|\V_{t-1}) = \mathbb{E}_{t-1}[\delta_j(\VV_t^i)]$, where $\delta_j(i)$ is the Kronecker delta function. In Section~\ref{subsec:vfoa_dynamics} we showed that there are up to 15 different possibilities for the VFOA transition probability. This enables us to derive an explicit formula for each case, see %appendix~B.
appendix~\ref{app:vfoa-learning}.
Consider for example one of these cases, namely $ \pilkl=P(\VV_t^i=l|\VV_{t-1}^i=k, \VV_{t-1}^k=l)$, which is the conditional probability that at $t$ person $i$ looks at target $l$, given that at $t-1$ person $i$ looked at person $k$ and that person $k$ looked at target $l$. This probability can be estimated with the following formula:
\begin{align*}
 \pilklhat =&\frac{\displaystyle\sum_{q=1}^{Q}\sum_{i=2}^{N_q}\sum_{t=2}^{T_q}\sum_{\substack{k=1\\k\neq i}}^{N_q}\sum_{l\neq i,k} \delta_l(\VV_t^{q,i}) \delta_k(\VV_{t-1}^{q,i}) \delta_l(\VV_{t-1}^{q,k})}{\displaystyle\sum_{q=1}^{Q}\sum_{i=2}^{N_q}\sum_{t=2}^{T_q}\sum_{\substack{k=1\\k\neq i}}^{N_q}\sum_{l\neq i,k} \delta_k(\VV_{t-1}^{q,i}) \delta_l(\VV_{t-1}^{q,k})}
  \end{align*}

\subsection{Learning the Gaussian Parameters}
\label{subsec:learning-normal}
% \textcolor{red}{Reformulate}

% The values for the transition and emission matrices $\A$ and $\C$ are entirely determined by the model definition (given the hyper-parameters $\alphabf$ and $\betabf$). However, the covariance of the noise during these processes ($\Gammabf$ and $\Sigmabf$) have no inherent reason to be fixed. We propose a EM-algorithm to train them. While the inference process requires to evaluate in parallel several possible values for each VFOA in a latent switching state fashion, the training is much easier as the VFOA has been annotated already. The graphical model in \ref{fig:switching_model} can be simplified as a simple Kalman filter in which the transition dynamic $\A_t$, $\bb_t$ is given, but varies over time. In this case, the classical forward-backward algorithm, using \textit{Rauch-Tung-Striebel}(RTS) equations, shall be used. All results presented here are from a bayesian point of view (see also~\cite{Bishop2006}).

In Section \ref{sec:inference} we described the derivation of the proposed model that is based on SKF. This model requires the parameters (means and covariances) of the Gaussian distributions defined in \eqref{eq:distribution_HL} and \eqref{eq:distribution_L}. Notice however that the mean \eqref{eq:expectation_HL} of \eqref{eq:distribution_HL} is parameterized by $\alphabf$. Similarly, the mean \eqref{eq:expectation_L} of  \eqref{eq:distribution_L} is parameterized by $\betabf$. Consequently, the model parameters are:
\begin{align}
\label{eq:parameters-theta}
  \thetabf &= (\alphabf,\betabf,\Gammabf_{\LL},\Sigmabf_{\H}),
\end{align}
and we remind that $\alphabf$ and $\betabf$ are $2\times 2$ diagonal matrices, $\Gammabf_{\LL}$  is a $8 \times 8$ covariance and and $\Sigmabf_{\H}$ is a $2\times 2$ covariance, and that we assumed that these matrices are common to all the active targets. Hence the total number of parameters is equal to $2+2+36+3=43$.

In the general case of SKF models, the discrete variables are unobserved both for learning and for inference. Here we propose a learning algorithm that takes advantage of the fact that the discrete variables, \ie VFOAs, are observed during the learning process, namely the VFOAs are annotated.
%While the inference process requires to evaluate in parallel several possible values for each VFOA in a latent switching state fashion, the training is much easier if the VFOAs have been already annotated. The graphical model in figure~\ref{fig:complete_model} can be simplified as a simple Kalman filter in which the transition dynamics $\A_t$, $\bb_t$ is provided, but varies over time.
We propose an EM algorithm adapted from~\cite{Bishop2006}.
% \textcolor{red}{Ok}
%The EM-algorithm is a sequential procedure that updates model parameters given data, up to convergence.
In the case of a standard Kalman filter, an EM iteration alternates between a forward-backward pass to compute the expected latent variables (E-step), and between the maximization of the expected complete-data log-likelihood (M-step).

We start by describing the M-step.
The complete-data log-likelihood is:
\begin{align}
   \ln P(&\LL^1, \H^1, \ldots, \LL^Q, \H^Q | \thetabf) \nonumber\\
  &= \sum_{q=1}^Q\sum_{i=2}^{N_q}\sum_{t=2}^{T_q}\ln P(\LL_t^{q,i}|\LL_{t-1}^{q,i}, \betabf, \Gammabf_\LL)\nonumber\\
  % =& \sum_{q=1}^Q \ln P(\LL_1^k|\PP_0) + \nonumber\\
    &  %\nonumber\\
   + \sum_{q=1}^Q\sum_{i=2}^{N_q}\sum_{t=1}^{T_q}\ln P(\H_t^{q,i}|\LL_t^{q,i}, \alphabf,  \Sigmabf_\H).\label{eq:c_data_log_likelihood}
\end{align}
By taking the expectation \wrt the posterior distribution $P(\LL^1,\ldots,\LL^Q | \H^1,\ldots,\H^Q, \thetabf)$, we obtain:
\begin{align}
  Q(\thetabf, \thetabf^{\text{old}}) &= \mathbb{E}_{\LL^1,\ldots,\LL^K|\thetabf^{\text{old}}}\left[\ln P(\LL^1, \H^1, \ldots, \LL^Q, \H^Q | \thetabf)\right],\label{eq:complete_data_log_likelihood}
\end{align}
which can be maximized \wrt to the parameters $\thetabf$, which yields closed-form expressions for the covariance matrices:
% \begin{align}
%   \PP_0 &= \frac{1}{Q} \sum_{q=1}^Q \mathbb{E}[\LL_1^q\LL_1^{q^\top}] - \mathbb{E}[\LL_1^q]\mathbb{E}[\LL_1^{q^\top}]\label{eq:update_P0}
% \end{align}
\begin{align}
  \Gammabf_\LL &= \frac{\displaystyle\sum_{q=1}^Q\sum_{i=2}^{N_q} \sum_{t=2}^{T_q} \mathbb{E}[(\LL_t^{q,i} - \mubf_{\LL,t}^{q,ij})(\LL_t^i - \mubf_{\LL,t}^{q,ij})^\top]}{\displaystyle\sum_{q=1}^Q (N_q-1)(T_q-1)}\label{eq:update_Gamma}
\end{align}
where $\mubf_{\LL,t}^{q,ij} = \A_t^{q,ij}\LL_{t-1}^{q,i}+\bb_t^{q,ij}$, \ie ~\eqref{eq:expectation_L}, and:
\begin{align}
  \Sigmabf_\H &= \frac{\displaystyle\sum_{q=1}^Q \sum_{i=2}^{N_q}\sum_{t=1}^{T_q} \mathbb{E}[(\H_t^{q,i}-\mubf_{\H,t}^{q,i})(\H_t^{q,i}-\mubf_{\H,t}^{q,i})^\top]}{\displaystyle\sum_{q=1}^Q (N_q-1)T_q}, \label{eq:update_Sigma}
\end{align}
where $\mubf_{\H,t}^{q,i}=\C\LL_t^{q,i}$, \ie \eqref{eq:expectation_HL}.

The estimation of $\alphabf$ and of $\betabf$ is carried out in the following way. $\partial Q(\thetabf, \thetabf^{\text{old}}) / \partial \beta_{1} =0 $  and $\partial Q(\thetabf, \thetabf^{\text{old}}) / \partial \beta_{2} =0 $ yield a set of two linear equations in the two unknowns:
\begin{align}
%\left\{
  \begin{array}{lll}
    \displaystyle \sum_{q=1}^Q\sum_{i=2}^{N_q}\sum_{t=2}^{T_q} \mathbb{E} \left[ (\LL_t^{q,i}-\mubf_{\LL,t}^{q,ij})^\top\Gammabf_\LL^{-1} \frac{\partial}{\partial \beta_{1}} (\LL_t^{q,i}-\mubf_{\LL,t}^{q,ij}) \right] = 0, \\
    \displaystyle \sum_{q=1}^Q\sum_{i=2}^{N_q}\sum_{t=2}^{T_q} \mathbb{E} \left[ (\LL_t^{q,i}-\mubf_{\LL,t}^{q,ij})^\top\Gammabf_\LL^{-1} \frac{\partial }{\partial \beta_{2}} (\LL_t^{q,i}-\mubf_{\LL,t}^{q,ij}) \right] = 0,
  \end{array}\label{eq:update_beta}
%\right.
\end{align}
and similarly:
\begin{align}
%\left\{
  \begin{array}{lll}
    \displaystyle \sum_{q=1}^Q\sum_{i=2}^{N_q}\sum_{t=1}^{T_q} \mathbb{E} \left[ (\H_t^{q,i}-\mubf_{\H,t}^{q,i})^\top\Sigmabf_\H^{-1} \frac{\partial}{\partial \alpha_{1}} (\H_t^{q,i}-\mubf_{\H,t}^{q,i}) \right] = 0, \\
    \displaystyle \sum_{q=1}^Q\sum_{i=2}^{N_q}\sum_{t=1}^{T_q} \mathbb{E} \left[ (\H_t^{q,i}-\mubf_{\H,t}^{q,i})^\top\Sigmabf_\H^{-1} \frac{\partial}{\partial \alpha_{2}} (\H_t^{q,i}-\mubf_{\H,t}^{q,i}) \right] = 0, \\
  \end{array}\label{eq:update_alpha}
%\right.
\end{align}
where as above, the expectation is taken \wrt to the posterior distribution.
%In equations~\eqref{eq:update_Gamma} to~\eqref{eq:update_alpha}, the expectation sign $\mathbb{E}$ must be understood as in equation~\eqref{eq:complete_data_log_likelihood}: the expectation \wrt the posterior distribution. All the components of $\thetabf$ have an update expression. This completes the M-step (M stands for Maximization).
Once the formulas above are expanded and once the means $\mubf_{\LL,t}^{q,ij}$ and $\mubf_{\H,t}^{q,i}$ are substituted with their expressions, the following terms remain to be estimated: $\mathbb{E}[\LL_t^{q,i}]$, $\mathbb{E}[\LL_t^{q,i}{\LL_t^{q,i}}^\top]$ and $\mathbb{E}[\LL_t^{q,i}{\LL_{t-1}^{q,i}}^\top]$.% rely on a previous estimate of all the parameters, and on the expectation on latent variable

The E-step provides estimates of these expectations via a forward-backward algorithm. For the sake of clarity, we drop the superscripts $i$ (active target index) and $q$ (recording index) up to equation~\eqref{eq:forward_backward_J} below. Introducing the notation $P(\LL_t | \H_1, \ldots, \H_t) = \mathcal{N}(\LL_t; \mubf_t, \PP_t)$, the forward-pass equations are:
\begin{align}
  \mubf_t   &= \A_t \mubf_{t-1} + \bb_t + \KK_t(\H_t - \C(\A_t \mubf_{t-1} + \bb_t))\\
  \PP_t     &= (\I-\KK_t\C)\PP_{t,t-1},
\end{align}
where:
\begin{align}
  \PP_{t,t-1} &= \A_t \PP_{t-1} \A_t^\top + \Gammabf_\LL,\\
  \KK_t     &= \PP_{t,t-1}\C^\top(\C\PP_{t,t-1}\C^\top + \Sigmabf_\H)^{-1}.
\end{align}
The backward pass estimates $P(\LL_t | \H_1, \ldots, \H_T) = \mathcal{N}(\LL_t; \hat{\mubf}_t, \hat{\PP}_t)$ and leads to
\begin{align}
  \hat{\mubf_t}   &= \mubf_t + \J_t (\hat{\mubf}_{t+1} - (\A_{t+1} \mubf_t + \bb_{t+1})),\\
  \hat{\PP_t}     &= \PP_t + \J_t (\hat{\PP}_{t+1} - \PP_{t+1,t}) \J_t^\top,
\end{align}
where:
\begin{align}
  \J_t &= \PP_t\A_{t+1}^\top(\PP_{t+1,t})^{-1}.\label{eq:forward_backward_J}
\end{align}
The expectations are estimated by performing a forward-backward pass over all the persons and all the recordings of the training data. This yields the following formulas:
\begin{align}
  \mathbb{E}[\LL_t^{q,i}] &= \hat{\mubf}_t^{q,i}\label{eq:expectation_li}\\
  \mathbb{E}[\LL_t^{q,i}{\LL_t^{q,i}}^\top] &= \hat{\PP}_t^{q,i} + \hat{\mubf}_t^{q,i} \mbox{$\hat{\mubf}^{q,i}_t$}^\top\label{eq:expectation_li_lit}\\
  \mathbb{E}[\LL_t^{q,i}{\LL_{t-1}^{q,i}}^\top] &= \hat{\PP}_t^{q,i}{\J_{t-1}^{q,i}}^\top + \hat{\mubf}_t^{q,i} \mbox{$\hat{\mubf}_{t-1}^{q,i}$}^\top\label{eq:expectation_li_litm1}
\end{align}

%% file: section_implementation.tex
\section{Implementation Details}
\label{sec:implementation}

 The proposed method was evaluated on the \emph{Vernissage} dataset~\cite{Jayagopi2012} and on the \emph{Looking At Each Other (LAEO)} dataset~\cite{Marin-Jimenez2014}. We describe in detail these datasets and their annotations. We provide implementation details and we analyse the complexity of the proposed algorithm.
% a total of twenty persons. Each recording lasts approximately ten minutes and contains a situated dialog between two persons and a robot. During the interaction the participants may gaze to each other, to the robot, or to some wall paintings. The dataset is recorded both with a motion capture system (a network of infrared cameras) and with a camera mounted onto the robot head. The motion capture device provides accurate head pose measurements for each participant in the dataset. The ground-truth VFOAs of all the participants were carefully annotated for each frame. Therefore, the \emph{Vernissage} dataset allows quantitative evaluation and benchmarking of VFOA estimation in a multi-party interaction scenario.
 %This section proposes a description of both datasets, and a few algorithmic details of the method.

\subsection{The \emph{Vernissage} Dataset}
\label{sec:vernissage-dataset}

\begin{figure}[h!]
\centering
%\subfigure[]{\includegraphics[width=0.6\linewidth]{back_view} \label{subfig:back-view} }
%\subfigure[]{\includegraphics[width=0.35\linewidth]{top_view} \label{subfig:top-view}}
\includegraphics[width=0.6\linewidth]{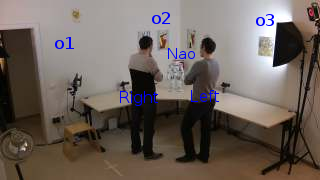} \label{subfig:back-view}
\includegraphics[width=0.35\linewidth]{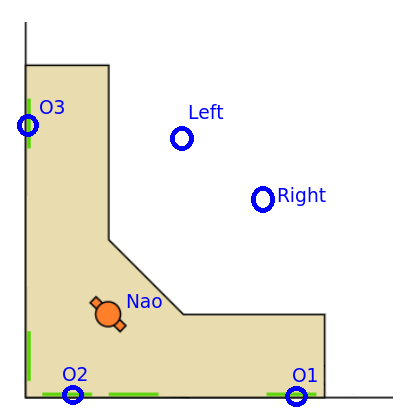} \label{subfig:top-view}
\caption{The \emph{Vernissage} setup.
Left: Global view of an ``exhibition" showing wall paintings, two participants, \ie left-p and right-p, and the NAO robot.
Right: Top view of the room showing the \emph{Vernissage} layout.}
\label{fig:back_view}
\end{figure}

%% Vernissage settings
The \emph{Vernissage} scenario can be briefly described as follows, \eg Fig.~\ref{fig:back_view}: there are three wall paintings, namely the passive targets denoted with $o_1$, $o_2$, and $o_3$ ($M=3$);  two persons, denoted left person (left-p) and right person (right-p), interact with the robot, hence $N=3$. The robot plays the role of an art guide, describing the paintings and asking questions to the two persons in front of him. Each recording is split into two roughly equal parts. The first part is dedicated to painting explanation, with a one-way interaction. The second part consists of a quiz, thus illustrating a dialog between the two participants and the robot, most of the time concerning the paintings.

%% Observations X and H
The scene was recorded with a camera embedded into the robot head and with a VICON motion capture system consisting of a network of infrared cameras, placed onto the walls, and of optical markers, placed onto the robot and people heads.  Both were recorded at 25 frames per second (fps). There is a total of ten recordings, each lasting ten minutes.
The VICON system provided accurate estimates of head positions, $\xoverline{\X}_{1:T}$ and head orientations, $\xoverline{\H}_{1:T}$. Head positions and head orientations were also estimated using from the RGB images gathered with the camera embedded into the robot head. The RGB images are processed as follows.
We use the OpenCV version of \cite{Viola2001} to detect faces and their bounding boxes which are then tracked over time using \cite{Bae2014}. Next, we extract HOG descriptors from each bounding box and apply a head-pose estimator, \eg \cite{Drouard2017}. This yields $\tilde{\H}_{1:T}$. The 3D head positions, $\tilde{\X}_{1:T}$, can be estimated using the line of sight through the face center and the bounding-box size, which provides a rough estimate of the depth along the line of sight.

In the remaining of this paper,  $\xoverline{\X}_{1:T}$ and $\xoverline{\H}_{1:T}$ are referred to as~\textit{Vicon Data};  $\tilde{\X}_{1:T}$ and $\tilde{\H}_{1:T}$ as~\textit{RGB Data}.
Because the whole setup was carefully calibrated, both Vicon and RGB Data are represented in the same coordinate frame.
%In parallel, for the algorithm to be tested in real conditions, a computer vision method can provide estimations for people head position and orientation.
%we used~\cite{Viola2001} to detect heads into the images from the robot.
%Bounding boxes for the faces have been manually annotated once every second of video; this provides a mean to evaluate the quality of a head detector. %Using these detections along with Nao proprioceptive information (position and orientation into the global system) allows to have an estimation of people's head position $\tilde{\H}_{1:T}$. On the other hand, we used~\cite{Drouard2016} to estimate the head orientation $\tilde{\X}_{1:T}$ from the head images. We evaluated our method both with accurate inputs $(\xoverline{\H}_{1:T}, \xoverline{\X}_{1:T})$ and with estimation from images $(\tilde{\H}_{1:T}, \tilde{\X}_{1:T})$.
%% Visual Focus of Attention V

In all our experiments we assumed that the passive targets are static and their positions are provided in advance.
The position of the robot itself is also known in advance and one can easily estimate the orientation of the robot head from motor readings.
Finally, the VFOAs of the participants were manually annotated in all the frames of all the recordings.

%In addition to the $N+M$ possible targets $\VV_t^i=j$ and the possibility of gaze aversion $\VV_t^i=0$, there is a third set of annotations, namely $\VV_t^i=unknown$. This annotation is used to indicate when the participants leave the field of view of the robot, or when the head pose algorithm fails.
%We now describe the acquisition of the RGB Data which relies on images from the camera.% uses the camera mounted onto the robot head.

%% The RGB images are processed as follows.
%% We use the OpenCV version of \cite{Viola2001} to detect faces and their bounding boxes which are then tracked over time using \cite{Bae2014}. Next, we extract HOG descriptors from each bounding box and apply a head-pose estimator, \eg \cite{Drouard2017}. This yields $\tilde{\H}_{1:T}$. The 3D head positions, $\tilde{\X}_{1:T}$, can be estimated using the line of sight through the face center and the bounding-box size, which provides a rough estimate of the depth along the line of sight.

\subsection{The \emph{LAEO} Dataset}
\label{sec:laeo-dataset}
  The \emph{LAEO} dataset  \cite{Marin-Jimenez2014} is an extension of the \emph{TVHID (TV Human Interaction Dataset)}~\cite{PatronPerez10}. It consists of 300 videos extracted from TV shows. At least two actors appear in each video engaged in four human-human interactions: handshake, highfive, hug, and kiss. There are 50 videos for each interaction and 100 videos with no interaction. The videos have been grabbed at 25~fps and each video lasts from five seconds to twenty-five seconds. \emph{LAEO} is further annotated, namely some of these videos are split into shots which are separated by cuts. There are 443 shots in total which are manually annotated whenever two persons look at each other, \cite{Marin-Jimenez2014}.
%The videos have been released at 25~fps and last between one and twenty-five seconds. Some of the original videos are composed of several shots, separated by cuts. \emph{LAEO}~\cite{Marin-Jimenez2014} introduces two novelties. First, the frames of the cuts are annotated. The original videos are separated into a total 443 shots. Second, each shot is manually annotated whether at least one pair of people are looking at each other or not during the sequence.

%\begin{figure}[h!]
%\centering
%\includegraphics[width=0.32\linewidth]{negative_0050_0054} \label{subfig:laeo1}
%\includegraphics[width=0.32\linewidth]{handshake_0036_0063} \label{subfig:laeo2}
%\includegraphics[width=0.32\linewidth]{handshake_0036_0078} \label{subfig:laeo3}
%\caption{Samples from three different shots from the \emph{LAEO}  dataset.}
%\label{fig:laeo_examples}
%\end{figure}

While there is no passive target in this dataset ($M=0$), the number of active targets ($N$) corresponds to the number of persons in each shot. In practice $N$ varies from one to eight persons.
All the faces in the dataset are annotated with a bounding box and with a coarse head-orientation label: frontal-right, frontal-left, profile-right, profile-left, backward. As with \emph{Vernissage}, we use the bounding-box center and size to estimate the 3D coordinates of the heads, $\xoverline{\X}_{1:T}$. We also assigned a yaw value to each one of the five coarse head orientations, $\xoverline{\H}_{1:T}$. We also computed
finer head orientations, $\tilde{\H}_{1:T}$, using \cite{Drouard2017}.
%There are no passive targets in any video ($M=0$); the number of active targets $N$ is the number of visible persons in the current shot, and varies from one to eight. Unlike \emph{Vernissage}, the camera is not carried by a robot: the position of the camera is neither an active nor a passive target. At each frame, the visible persons are annotated with a head bounding box and a 5-classes head yaw orientation: frontal-right, frontal-left, profile-right, profile-left, backwards. From this, it is possible to estimate the 3D head positions $\xoverline{\X}_{1:T}$ relatively to the camera. A reliable yet inaccurate estimate of the head orientations $\xoverline{\H}_{1:T}$ were set by attributing actual angles for each yaw annotation. Continuous head orientations $\tilde{\H}_{1:T}$ were also estimated using~\cite{Baltrusaitis2016} \textcolor{red}{(How about Vincent's method ~\cite{Drouard2017}?)}.

\subsection{Algorithmic Details}
\label{subsec:implementation}

The inference procedure is summarized in Algorithm~\ref{algo:inference}. This is basically an iterative filtering procedure. The update step consists of applying the recursive relationship, derived in Section~\ref{sec:inference}, to $\mubf_t^{ij}$, $\Sigmabf_t^{ij}$ and $c_t^{ij}$, using $\mubf_t^{ijk}$, $\Sigmabf_t^{ijk}$ and $c_{t-1,t}^{ijk}$ as intermediate variables. The VFOA is chosen using MAP, given observations up to the current frame, and the gaze direction is the mean of the filtered distribution (the first two components of $\mubf_t^{ij}$ are indeed the mean for the pan and tilt gaze angles). % Let's detail two problems that were not addressed until now : How to initialize the recursion, and how the observations are obtained.

\begin{algorithm}
  \caption{Inference}\label{algo:inference}
  \begin{algorithmic}[1]
    \Procedure{GazeAndVFOA}{}
    \State $\X_1, \H_1 \gets \Call{GetObservations}{time=1}$
    \State $c_1, \mubf_1, \Sigmabf_1 \gets \Call{Initialization}{\H_1,\X_1}$
    \State $\VV_1^i \gets\argmax_j c_1^{ij}$
    \State $\G_1^i  \gets \mubf_1^{ij}[1..2]$
    \For{$t=2..T$}
    \State $\X_t, \H_t \gets \Call{GetObservations}{time=t}$
    \State $c_t, \mubf_t, \Sigmabf_t \gets \Call{Update}{\H_t,\X_t, c_{t-1}, \mubf_{t-1}, \Sigmabf_{t-1}}$
    \State $\VV_t^i \gets\argmax_j c_t^{ij}$
    \State $\G_t^i  \gets \mubf_t^{ij}[1..2]$
    \EndFor
    \State \Return $\V_{1:T}, \G_{1:T}$
    \EndProcedure
  \end{algorithmic}
\end{algorithm}

Let's now describe the initialization procedure used by Algorithm~\ref{algo:init}. In a probabilistic framework, parameter intialization is generally addressed by defining an initial distribution, \eg $P(\LL_1|\V_1)$. Here, we did not explicitly define such a distribution. Initialization is based on the fact that, with repeated similar observation inputs, the algorithm reaches a steady-state. The initialization algorithm uses a repeated update method with initial observation to provide an estimate of gaze and of reference directions. Consequently, the initial filtering distribution $P(\LL_1, \V_1 | \H_1, \X_1)$ is implicitly defined as the expected stationary state.
%In the case of \emph{Vernissage} sequences, this problem is not central as it concerns one frame out of thousands (25 per second, videos last several minutes). Moreover, there are only a handful of videos, limiting the relevance of training such a distribution. For these reasons, the filtering distribution is initialized as follow: $P(\LL_1^i|\VV_1^i=j, \H_1, \X_1) = \mathcal{N}(\LL_1^i; \mu_1^{ij}, \Sigmabf_1^{ij})$ where $\mu_1^{ij}$ and $\Sigmabf_1^{ij}$ are obtained respectively from $[\H_1^i;\zero;\H_1^i;\zero]$ and $\Gammabf_\LL$ by taking a few filtering iterations using $\H_1$ and $\X_1$ as inputs.

% \begin{algorithm}
%   \caption{Update}\label{algo:update}
%   \begin{algorithmic}[1]
%     \Procedure{Update}{$\H_t,\X_t, c_{t-1}, \mubf_{t-1}, \Sigmabf_{t-1}$}
%     \State Compute $c_{t-1,t}^{ijk}$ from~\eqref{eq:cijk}
%     \State $\mubf_t^{ijk},\Sigmabf_t^{ijk} \gets \Call{Filter}{\mubf_{t-1}^{ik}, \Sigmabf_{t-1}^{ik}, \A_t^{ij}, \bb_t^{ij}}$
%     \State Compute $\mubf_t^{ij}$ and $\Sigmabf_t^{ij}$ from~\eqref{eq:update_mutij} and~\eqref{eq:update_sigmatij}
%     \State $c_t^{ij} \gets \sum_k c_{t-1,t}^{ijk}$
%     \EndProcedure
%   \end{algorithmic}
% \end{algorithm}

\begin{algorithm}
  \caption{Initialization}\label{algo:init}
  \begin{algorithmic}[1]
    \Procedure{Initialization}{$\H_1,\X_1$}
    \State $\mubf_{in} \gets [\H_1; \zero; \H_1; \zero]$
    \State $\Sigmabf_{in} \gets \I$
    \State $c_{in} \gets \frac{1}{N+M} (\textit{Uniform})$
    \While {Not Convergence}
    \State $c_{in}, \mubf_{in}, \Sigmabf_{in} \gets \Call{Update}{\H_1,\X_1, c_{in}, \mubf_{in}, \Sigmabf_{in}}$
    \EndWhile
    \State \Return $c_{in}, \mubf_{in}, \Sigmabf_{in}$
    \EndProcedure
  \end{algorithmic}
\end{algorithm}

  \subsection{Algorithm Complexity}
  \label{subsec:complexity}
  The computational complexity of Algorithm~\ref{algo:inference} is
  \begin{align}
  \label{eq:complexity_total}
    C &= T (C_O+C_U) + T_I C_U,
  \end{align}
where $T$ is the number of frames in a test video, $T_I$ is the number of iterations needed by the Algorithm~\ref{algo:init} (initialization) to converge, $C_O$ is the computational complexity of \textsc{GetObservation} and $C_U$ is the computational complexity of \textsc{Update}.
%Since the proposed algorithm is online, we mainly are interested in the cost of one loop iteration; as long as the computational time required is lower than the frame duration, the computation will not be delayed.
The complexity  of one iteration of Algorithm~\ref{algo:inference} is $C_O+C_U$. $C_O$ depends on face detection and head pose estimation algorithms. Hence we concentrate on $C_U$. From Section~\ref{sec:inference} one sees that the following values need to be computed: $P(\H_t^i|\VV_{t}^i=j, \VV_{t-1}^i=k, \H_{1:t-1}, \X_{1:t-1})$, $c_{t-1,t}^{ijk}$, $\mubf_t^{ijk}$, $\Sigmabf_t^{ijk}$, and then  $c_t^{ij}$, $\mubf_t^{ij}$ and $\Sigmabf_t^{ij}$, for each active target $i$, and for each combination of targets $j$ and $k$ different from $i$. There are $N$ possible values for $i$ and $(N+M)$ possible values for $j$ and $k$. Then,
  \begin{align}
    C_U = K \times N(N+M)^2,
    \label{eq:complexity_update}
  \end{align}
%is delegated to a third-party program but it cannot be dropped in the general formula since detecting people and computing head orientation require non-negligeable computing time (The Viola-Jones detector is easily made real-time but most head orientation estimators require very high computational power to be real-time). Finally, let's analyse the complexity $C_U$. For one update, recall from section~\ref{sec:inference} that the following values need to be computed $P(\H_t^i|\VV_{t}^i=j, \VV_{t-1}^i=k, \H_{1:t-1}, \X_{1:t-1})$, $c_{t-1,t}^{ijk}$, $\mubf_t^{ijk}$, $\Sigmabf_t^{ijk}$, and then  $c_t^{ij}$, $\mubf_t^{ij}$ and $\Sigmabf_t^{ij}$, for each active target $i$, and for each combination of targets $j$ and $k$ different from $i$. There are $N$ possible values for $i$ and $(N+M)$ possible values for $j$ and $k$. Then,
%  \begin{align}
%    C_U &\propto N(N+M)^2.\label{eq:complexity_update}
%  \end{align}
where $K$ is a factor whose complexity can be estimated as follows. The most time-consuming part is the Kalman Filter algorithm used to estimate $\mubf_t^{ijk}$ and $\Sigmabf_t^{ijk}$ from $\mubf_t^{ik}$ and $\Sigmabf_t^{ik}$. These calculations are dominated by several 8$\times$8 and 2$\times$8 matrix inversions and multiplications. By neglecting scalar multiplications and matrix additions, and by denoting with $C_{KF}$ the complexity of the Kalman filter, we obtain that $K\approx C_{KF}$ and hence
$C_U \approx C_{KF} \times N(N+M)^2$.

%% file: section_experimental_results.tex
\section{Experimental results}
\label{sec:results}

%We start by explaining how the learning is carried out, then we show results obtained with the Vicon and RGB data. We compare our method with two other methods based on HMMs, followed by a short discussion.

\subsection{\emph{Vernissage} Dataset}

We applied the same experimental protocol to the Vicon and RGB data. We used a \textit{leave-one-video-out} strategy for training. The test is performed on the left out video.
We used the frame recognition rate (FRR) metrics to quantitatively evaluate the methods. FRR computes the percentage of frames for which the VFOA is correctly estimated. One should note however that the ground-truth VFOAs were obtained by manually annotating each frame in the data. This is subject to errors since the annotator has to associate a target with each person.
%This metric was used as well in~\cite{Ba2009,Sheikhi2012}. It is noteworthy that the \emph{Vernissage} dataset has a high class unbalance due to the nature of the scenario. The Nao robot attracts the attention of participants during a significant proportion of time (More than 40\%). For this reason, we tried different metrics to take this into account, for instance the F-score (a combination of precision and recall) or the Cohen's Kappa. However, in all our tests, these metrics were strongly correlated with the FRR.

The VFOA transition probabilities and the model parameters were estimated using the learning method described in Section~\ref{sec:learning}. Appendix~\ref{app:vfoa-learning} provides the formulas used for estimating the VFOA transition probabilities given the annotated data. Notice that the fifteen transitions probabilities thus estimated are identical for both data, Vicon and RGB.

The Gaussian parameters, \ie \eqref{eq:parameters-theta}, were estimated using the EM algorithm of Section~\ref{subsec:learning-normal}. This learning procedure requires head-pose estimates as well as the targets locations, estimated as just explained. Since these estimates are different for the two kinds of data (Vicon and RGB) we carried out the learning process twice, with the Vicon data and with the RGB data. The EM algorithm needs initialization. The initial parameter values for $\alphabf$ and $\betabf$ are
$\alphabf^0 = \betabf^0 = \diag (0.5, 0.5)$.
Matrices $\Sigmabf_{\H}$ and $\Gammabf_{\LL}$ defined in \eqref{eq:covariance_L} are initialized with isotropic covariances:
$\Sigmabf_{\H}^0 =\sigma \I_2$, $\Gammabf_{\G}^0 =\Gammabf_{\dot{\G}}^0=\gamma \I_2$, and $\Gammabf_{\R}^0=\Gammabf_{\dot{\R}}^0=\eta \I_2$ with $ \sigma=15$, $ \gamma =5$, and $\eta =0.5$. In particular, this initialization is consistent with  \eqref{eq:covariance-change}. In practice we noticed that the covariances estimated by training remain consistent with  \eqref{eq:covariance-change}.

%of the normal distributions are computed from a EM-algorithm and need to be initialized. $\alphabf$ and $\betabf$ are both initialized as $\diag (0.5, 0.5)$ and converge respectively to values around $\diag (0.65, 0.45)$ and $\diag (0.9, 0.95)$. For instance, for video 30, the training is done with video 09 to 27 and converge to $\alphabf=\diag(0.61,0.51)$ and $\betabf=\diag(0.91,0.96)$. The value of $\alphabf$ is consistent with previous work using similar head pose modelling \eg~\cite{Ba2009, Sheikhi2015}. The covariance matrices are first initialized as isotropic covariances as $\Sigmabf_{\H}=\sigma_{\H}^2\I_2$, $\Gammabf_{\G}=\Gammabf_{\dot{\G}}=\gamma_{\G}^2\I_2$ and $\Gammabf_{\R}=\Gammabf_{\dot{\R}}=\gamma_{\R}^2\I_2$ with $\sigma_{\H}=15^\circ$, $\gamma_{\G}=5^\circ$ and $\gamma_{\R}=0.5^\circ$. The training leads to dense covariance matrices.

\subsection{Results with Vicon Data}
%% \paragraph{RGB Data}

\begin{figure}[t!]
  \begin{center}
    \includegraphics[width=0.45\linewidth]{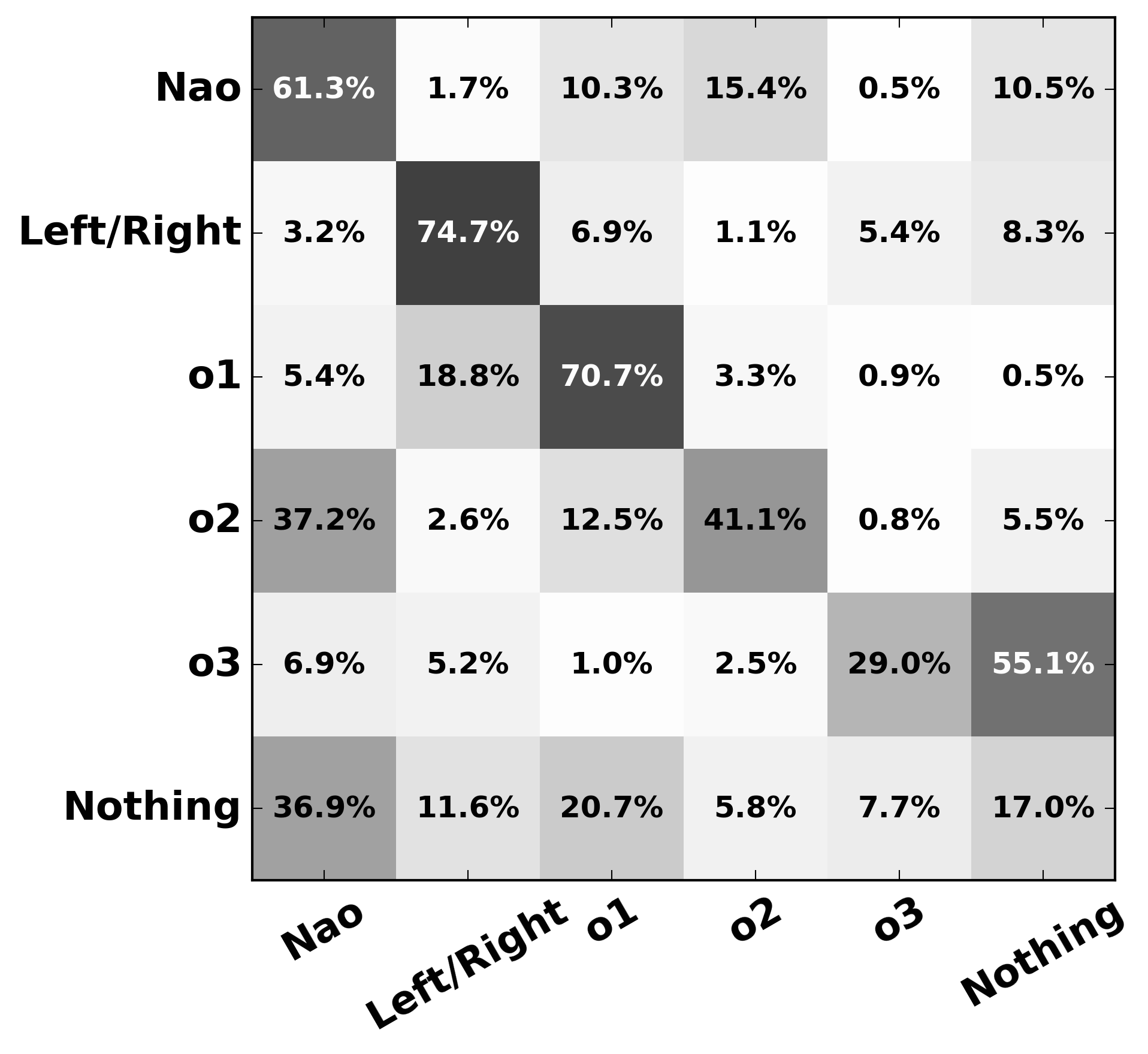}
    \includegraphics[width=0.45\linewidth]{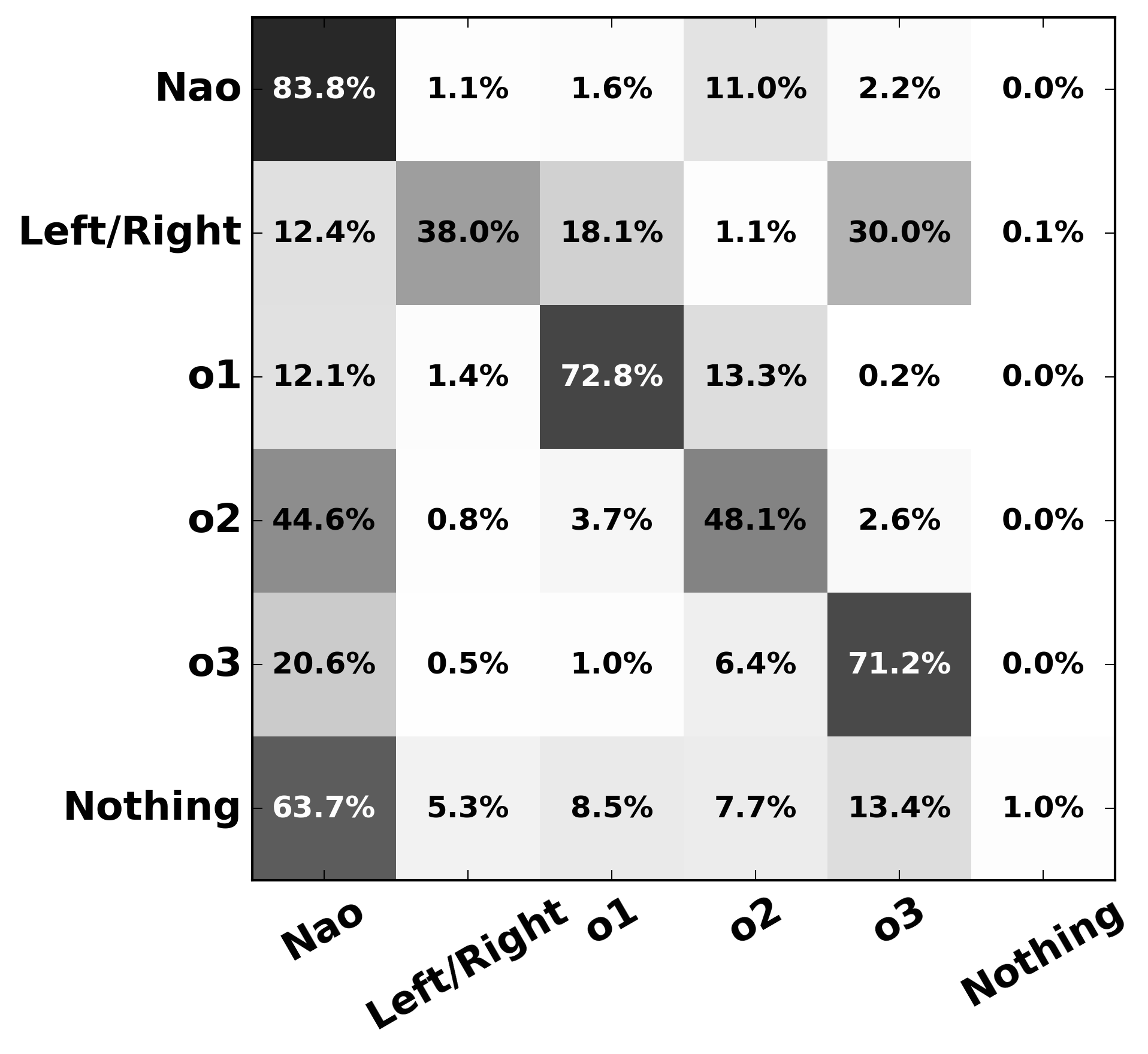}
  \end{center}
  \caption{Confusion matrices for the Vicon data. Left: \cite{Ba2009}. Right: Proposed algorithm.
    Row-wise: ground-truth VFOAs. Column-wise: estimated VFOAs. Diagonal terms represent the recall.
  }
  \label{fig:confusion_vicon}
\end{figure}

\begin{figure*}[t!]
\centering
\begin{minipage}[t]{.24\textwidth}
  \centering
  \includegraphics[width=  \linewidth]{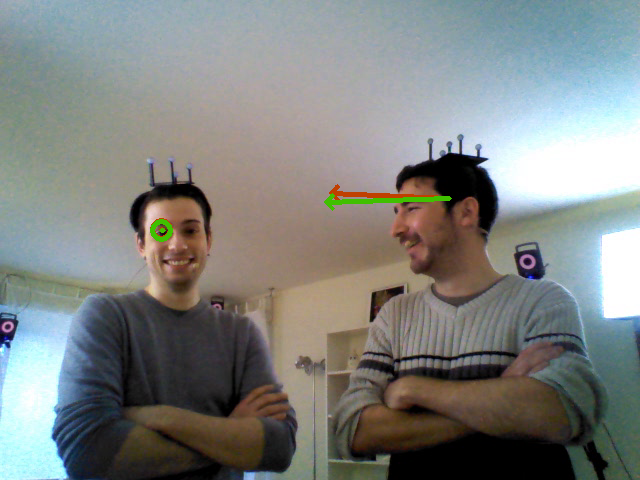}
  \includegraphics[width=.8\linewidth]{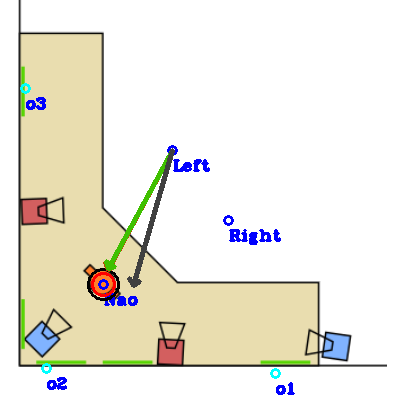}
  \includegraphics[width=.8\linewidth]{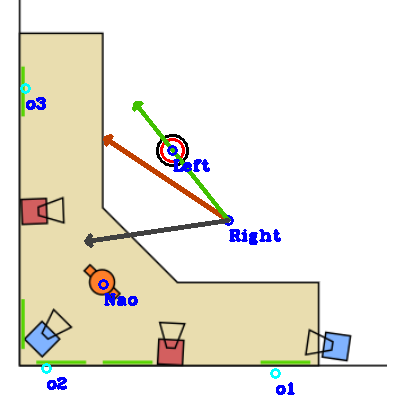}
\end{minipage}\hfill
\begin{minipage}[t]{.24\textwidth}
  \centering   \centering
  \includegraphics[width=  \linewidth]{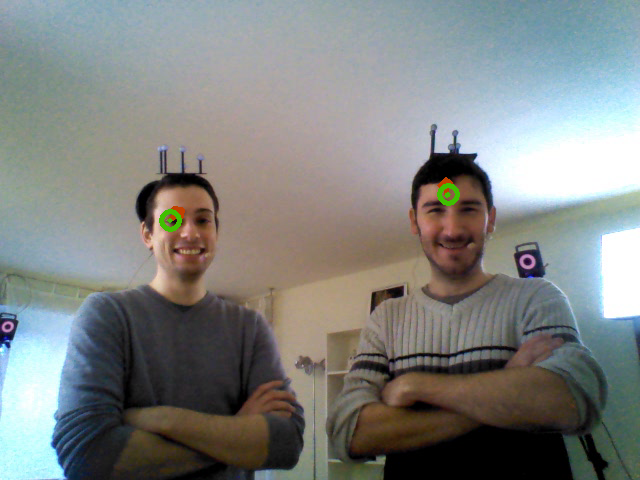}
  \includegraphics[width=.8\linewidth]{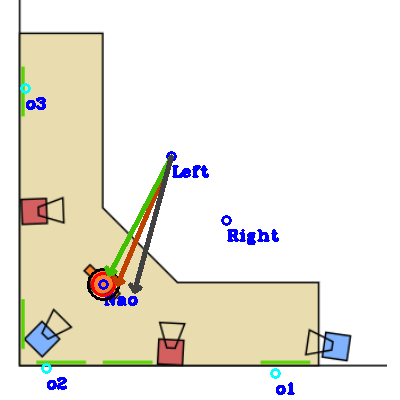}
  \includegraphics[width=.8\linewidth]{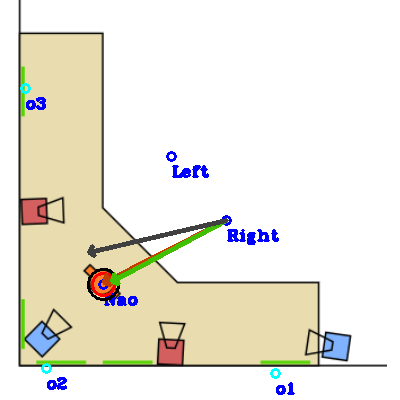}
\end{minipage}\hfill
\begin{minipage}[t]{.24\textwidth}
  \centering   \centering
  \includegraphics[width=  \linewidth]{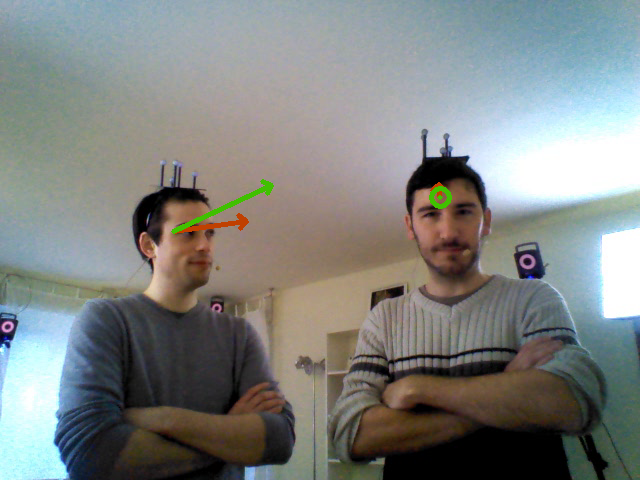}
  \includegraphics[width=.8\linewidth]{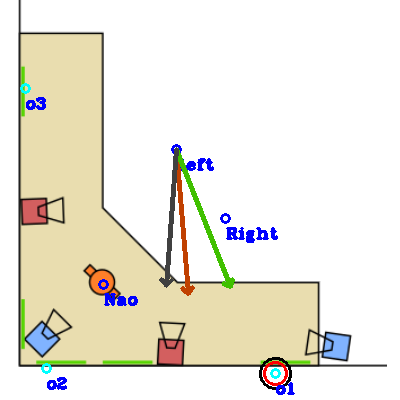}
  \includegraphics[width=.8\linewidth]{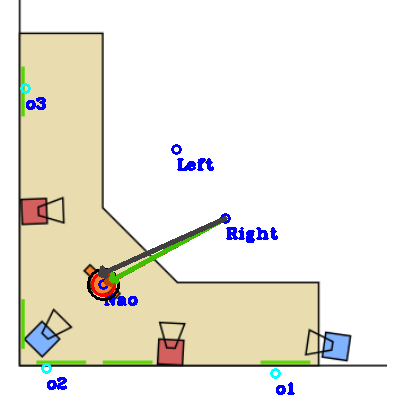}
\end{minipage}\hfill
\begin{minipage}[t]{.24\textwidth}
  \centering   \centering
  \includegraphics[width=  \linewidth]{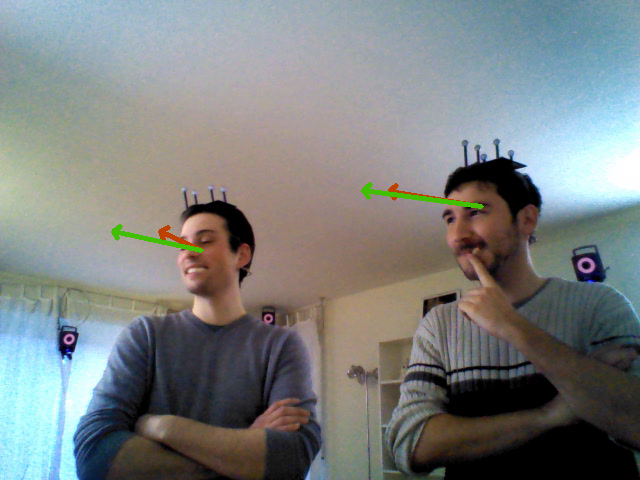}
  \includegraphics[width=.8\linewidth]{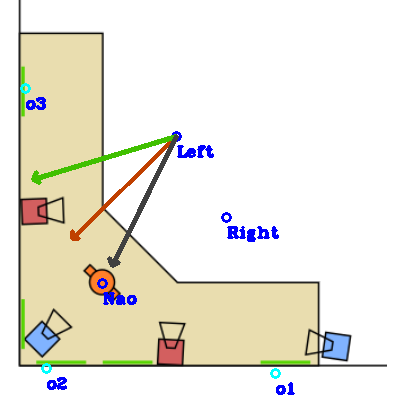}
  \includegraphics[width=.8\linewidth]{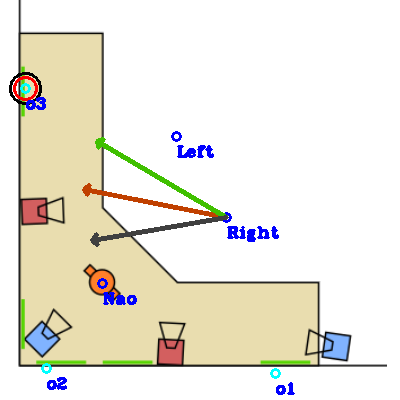}
\end{minipage}\hfill
   \caption{
     Results obtained with the proposed method on Vicon data. Gaze directions are shown with green arrows, head reference directions with dark-grey arrows and observed head directions with red arrows. The ground-truth VFOA is shown with a black circle. The top row displays the image of the robot-head camera. Top views of the room show results obtained for the left-p (middle row) and for the right-p (bottom row). In the last example the left-p gazes at ``nothing".
}
\label{fig:screenshots_vicon}
\end{figure*}

The FRR of the estimated VFOAs for the Vicon data are summarized in Table~\ref{table:res_vicon_vfoa}. A few examples are shown in Figure~\ref{fig:screenshots_vicon}.
\begin{table}[t!]
  \centering
  \caption{  \label{table:res_vicon_vfoa} FRR scores of the estimated VFOAs for the Vicon data for the left and right persons (left-p and right-p).}
  \begin{tabular}{|c|c|c|c|c|}
    \hline
    Recording &\multicolumn{2}{|c|}{Ba \& Odobez~\cite{Ba2009}}&\multicolumn{2}{|c|}{Proposed}\\
    \hline
    &left-p&right-p&left-p&right-p\\
    \hline
    09&    51.6 &\bf 65.1&\bf 59.8&    61.4\\
    \hline
    10&    64.3 &\bf 74.4&\bf 76.5&    65.0\\
    \hline
    12&    53.5 &\bf 67.6&\bf 61.6&    63.2\\
    \hline
    15&\bf 67.1 &    46.2&    64.8&\bf 67.6\\
    \hline
    18&    37.5 &    28.3&\bf 62.0&\bf 53.7\\
    \hline
    19&\bf 56.7 &    45.4&    54.5&\bf 60.4\\
    \hline
    24&    44.9 &    49.0&\bf 59.7&\bf 54.7\\
    \hline
    26&    40.3 &    32.9&\bf 43.6&\bf 43.1\\
    \hline
    27&    65.8 &    72.0&\bf 79.8&\bf 78.3\\
    \hline
    30&    69.1 &    49.1&\bf 72.0&\bf 63.9\\
    \hline
    Mean&\multicolumn{2}{|c|}{54.5}&\multicolumn{2}{|c|}{\bf 62.6}\\
    \hline
  \end{tabular}
\end{table}
%
% One of the reasons that explain the variability among the participants is the willingness to move the head.
The  FRR score varies between $28.3\%$ and $74.4\%$ for \cite{Ba2009} and between $43.1\%$ and $79.8\%$ for the proposed method. Notice that high scores are obtained by both methods for recording \#27. Similarly, low scores are obtained for recording \#26. Since both methods assume that head motions and gaze shifts occur synchronously, an explanation could be that this hypothesis is only valid for some of the participants.
The confusion matrices for VFOA classification using Vicon data are given in Figure~\ref{fig:confusion_vicon}.
There are a few similarities between the results obtained with the two methods. In particular, wall painting \#$o_2$ stands just behind Nao and both methods don't always discriminate between these two targets. In addition, the head of one of the persons is often aligned with painting \#$o_1$ from the viewpoint of the other person. A similar remark holds for painting \#$o_3$. As a consequence both methods often confuse the VFOA in these cases. This can be seen in the third image of Figure~\ref{fig:screenshots_vicon}. Indeed, it is difficult to estimate whether the left person (left-p) looks at \#$o_1$ or at \textit{right-p}.

Finally, both methods have problems with recognizing the VFOA ``nothing'' or gaze aversion ($\VV_t^i=0$). We propose the following explanation: the targets are widespread in the scene, hence it is likely that an acceptable target lies in most of the gaze directions. Moreover, Nao is centrally positioned, therefore the head orientation used to look at Nao is similar to the resting head orientation used for gaze aversion.
However, in \cite{Ba2009} the reference head orientation is fixed and poorly suited for dynamic head-to-gaze mapping, hence the high error rate on painting \#$o_3$. Our method favors the selection of a target, either active or passive, over the no target (nothing) case.

\subsection{Results with RGB Data}
%% \paragraph{RGB Data}
\label{subsec:RGBdata}
The RGB images were processed as described in section~\ref{sec:vernissage-dataset} above in order to obtain head orientations, $\tilde{\H}_{1:T}$, and 3D head positions,
$\tilde{\X}_{1:T}$. Table~\ref{table:bbox} shows the accuracy of these measurements (in degrees and in centimeters), when compared with the ground truth provided by the Vicon motion capture system. As it can be seen, while the head orientation estimates are quite accurate, the error in estimating the head positions can be as large as 0.8~m for participants lying in between 1.5~m and 2.5~m in front of a robot, \eg recordings \#19 and \#24. In particular this error increases as a participant is farther away from the robot. In these cases, the bounding box is larger than it should be and hence the head position is, on an average, one meter closer than the true position. These relatively large errors in 3D head position affect the overall behavior of the algorithm.

\begin{table}[b!]
  \centering
  \caption{Mean error for head pose estimations from RGB data, for the left person (left-p) and the right person (right-p). The errors in head position (centimeters) and orientation (degrees) are computed with respect to values provided by the motion capture system.}
  \label{table:bbox}
  \begin{tabular}{|c|c|c|c|c|c|c|}
    \hline
    Video &\multicolumn{2}{|c|}{Position error (cm)}&\multicolumn{2}{|c|}{Pan error}&\multicolumn{2}{|c|}{Tilt error}\\
    \hline
    &left-p&right-p&left-p&right-p&left-p&right-p\\
    \hline
    09&18.1&20.8&4.4\degree&4.8\degree&3.7\degree&3.2\degree\\
    \hline
    12&35.7&41.5&4.8\degree&5.5\degree&2.6\degree&3.8\degree\\
    \hline
    18&36.9&12.8&6.8\degree&3.7\degree&5.8\degree&2.5\degree\\
    \hline
    19&86.0&87.4&4.0\degree&5.8\degree&2.7\degree&3.7\degree\\
    \hline
    24&86.5&73.9&3.3\degree&3.5\degree&2.8\degree&2.7\degree\\
    \hline
    26&50.2&56.9&7.4\degree&9.0\degree&4.1\degree&5.2\degree\\
    \hline
    27&64.5&58.3&4.1\degree&5.8\degree&3.2\degree&4.4\degree\\
    \hline
    30&16.7&13.3&2.8\degree&2.9\degree&1.8\degree&2.7\degree\\
    \hline
    Mean&\multicolumn{2}{|c|}{46.4}&\multicolumn{2}{|c|}{5.0\degree}&\multicolumn{2}{|c|}{3.3\degree}\\
    \hline
  \end{tabular}
\end{table}

\begin{figure}[t]
  \begin{center}
    \includegraphics[width=0.48\linewidth]{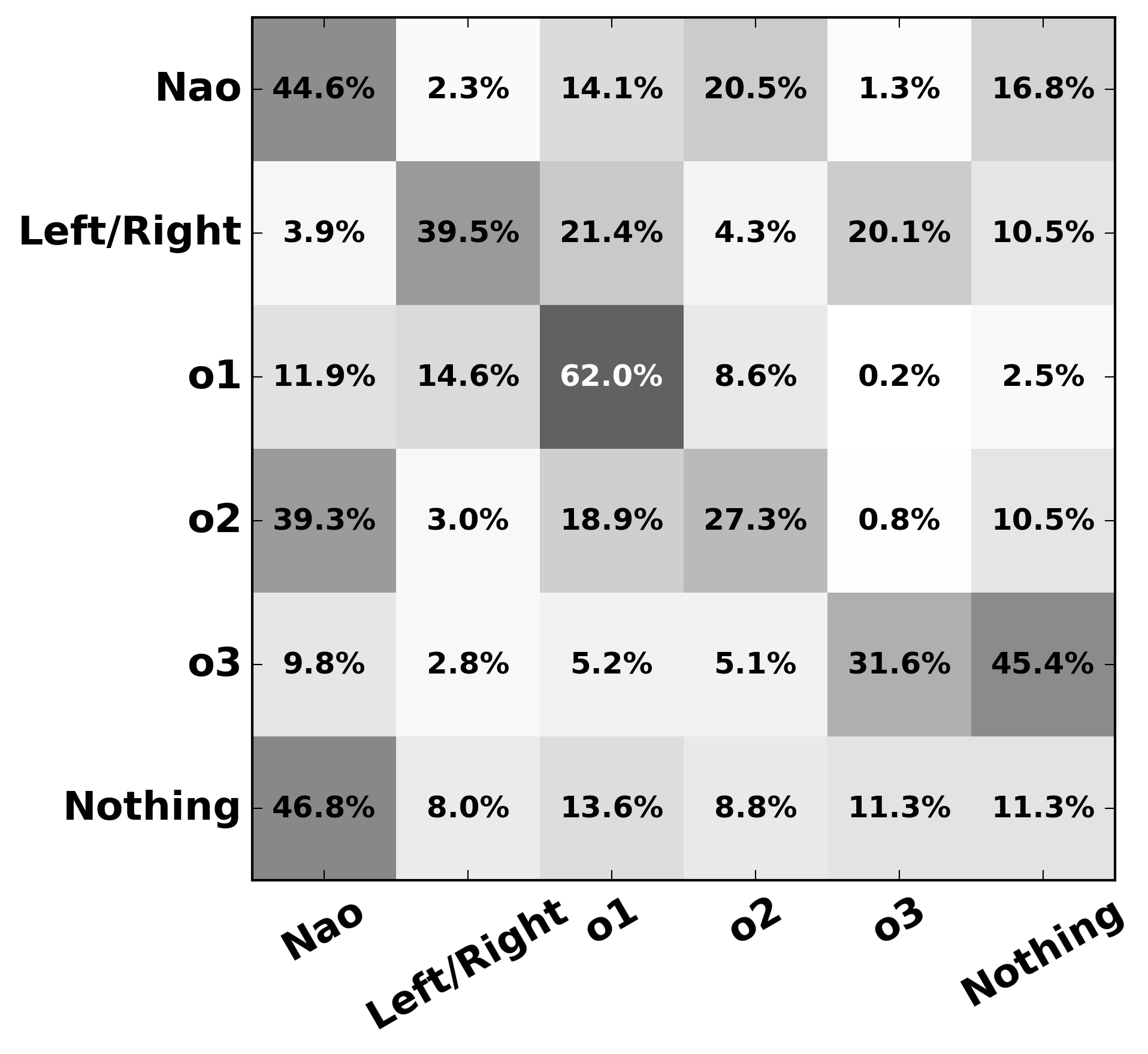}
    \includegraphics[width=0.48\linewidth]{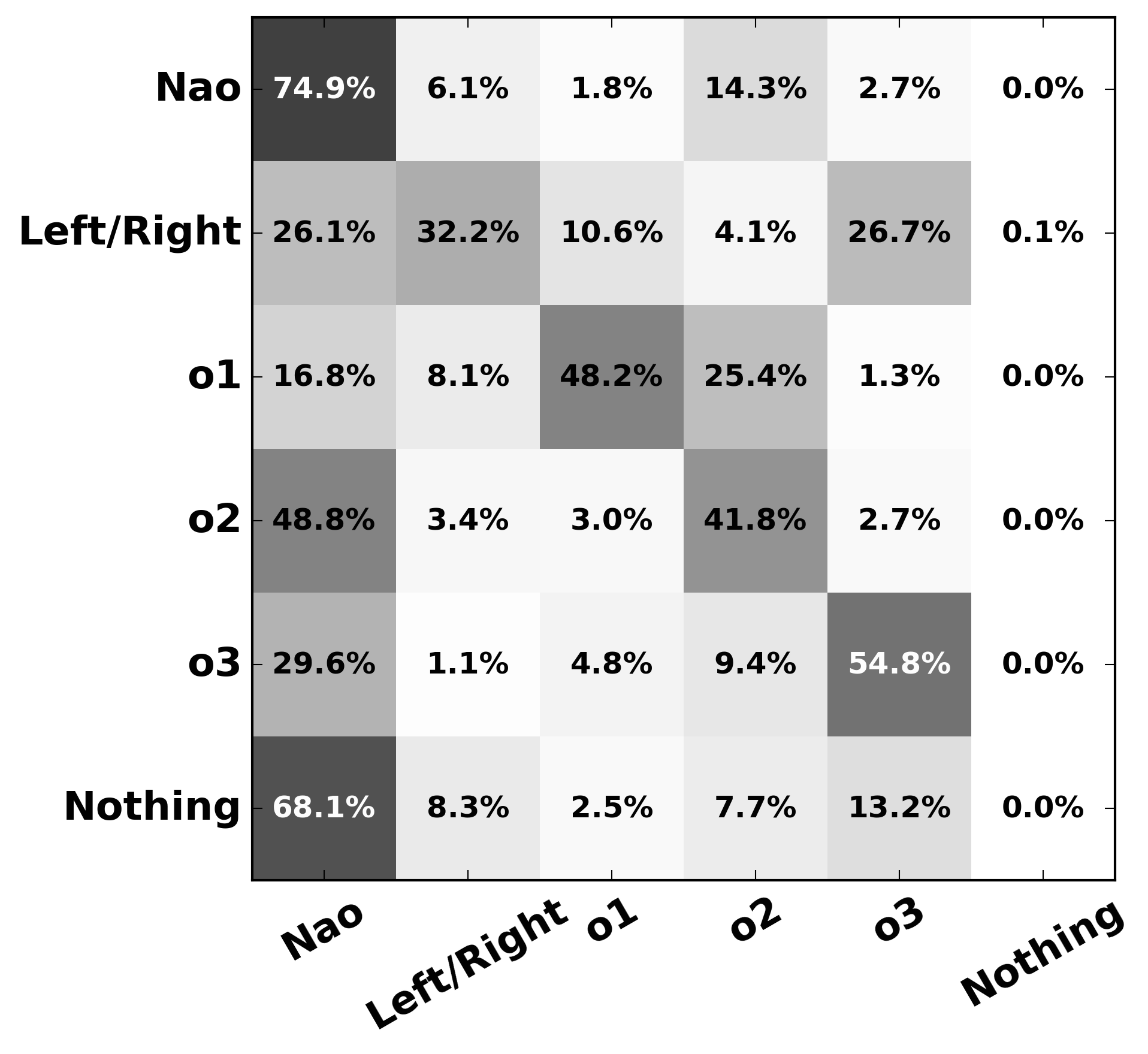}
  \end{center}
  \caption{Confusion matrices for the RGB data. Left: \cite{Ba2009}. Right: Proposed algorithm.
    Row-wise: ground-truth VFOAs. Column-wise: estimated VFOAs. Diagonal terms represent the recall.
  }
  \label{fig:confusion_visual}
\end{figure}

\begin{table}[b!]
  \centering
  \caption{FRR scores of the estimated VFOAs obtained with \cite{Ba2009} and with the proposed method for the RGB data. The last two columns show the 3D head position errors of Table~\ref{table:bbox}.}
  \label{table:res_tracked_vfoa}
  \begin{tabular}{|c|c|c|c|c|c|c|}
    \hline
    Video &\multicolumn{2}{|c|}{Ba \& Odobez~\cite{Ba2009}}&\multicolumn{2}{|c|}{Proposed}&\multicolumn{2}{|c|}{Head pos. error}\\
    \hline
    &left-p&right-p&left-p&right-p&left-p&right-p\\
    \hline
    09&    50.3&\bf 59.8&\bf 58.1&    55.9 &18.1&20.8 \\
    \hline
    12&    54.2&    14.8&\bf 59.0&\bf 46.5&35.7&41.5\\
    \hline
    18&    39.0&    16.1&\bf 64.2&\bf 33.1 &36.9&12.8 \\
    \hline
    27&    38.2&    17.1&\bf 53.3&\bf 55.1 &64.5&58.3 \\
    \hline
    30&\bf 61.6&    44.6&    54.7&\bf 66.6 &16.7&13.3\\
    \hline
    Mean&\multicolumn{2}{|c|}{39.0}&\multicolumn{2}{|c|}{\bf 54.7}&\multicolumn{2}{|c|}{}\\
    \hline
    % Vicon&\multicolumn{2}{|c|}{56.6}&\multicolumn{2}{|c|}{65.9}\\
    % \hline
  \end{tabular}
\end{table}
The FRR scores obtained with the RGB data are shown in Table~\ref{table:res_tracked_vfoa}.
As expected the loss in accuracy is correlated with the head position error: the results obtained with recordings \#09 and \#30 are close to the ones obtained with the Vicon data, whereas there is a significant loss in accuracy for the other recordings. The loss is notable for \cite{Ba2009} in the case of the right person (right-p) for the recordings \#12, \#18 and \#27. %This is easily explained by the fact that the chosen reference depends on the relative position of targets.
The confusion matrices obtained with the RGB data are shown on Fig.~\ref{fig:confusion_visual}.
%The patterns observed with the Vicon data are also present in RGB data experiments.

%
\begin{figure*}[t!]
\centering
\begin{minipage}[t]{.24\textwidth}
  \centering
  \includegraphics[width=  \linewidth]{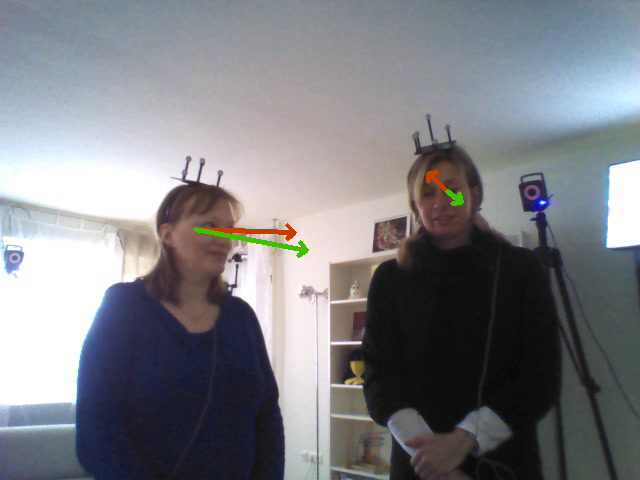}
  \includegraphics[width=.8\linewidth]{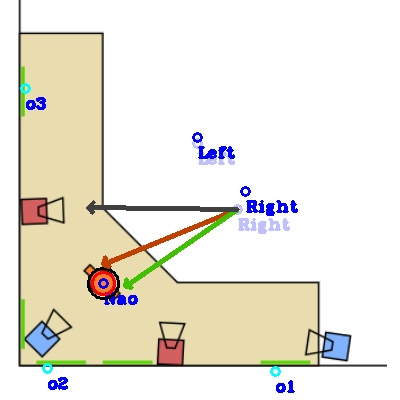}
  \includegraphics[width=.8\linewidth]{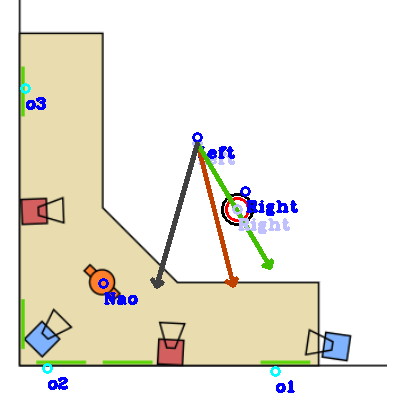}
\end{minipage}\hfill
\begin{minipage}[t]{.24\textwidth}
  \centering   \centering
  \includegraphics[width=  \linewidth]{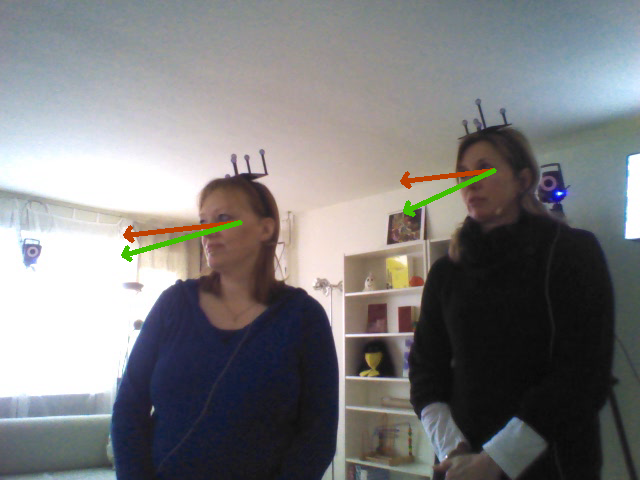}
  \includegraphics[width=.8\linewidth]{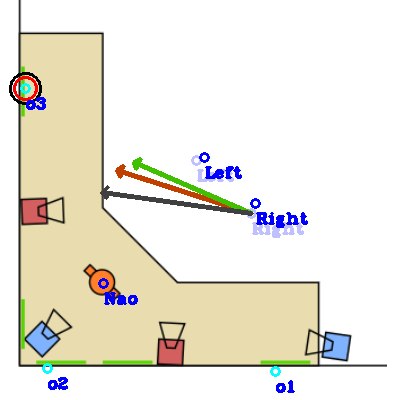}
  \includegraphics[width=.8\linewidth]{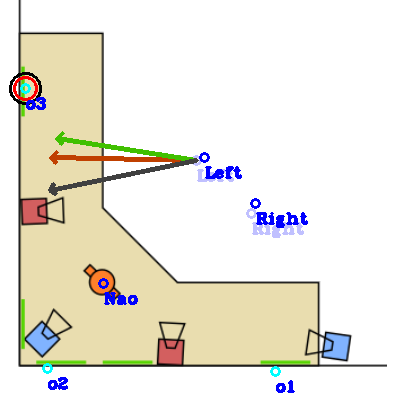}
\end{minipage}\hfill
\begin{minipage}[t]{.24\textwidth}
  \centering   \centering
  \includegraphics[width=  \linewidth]{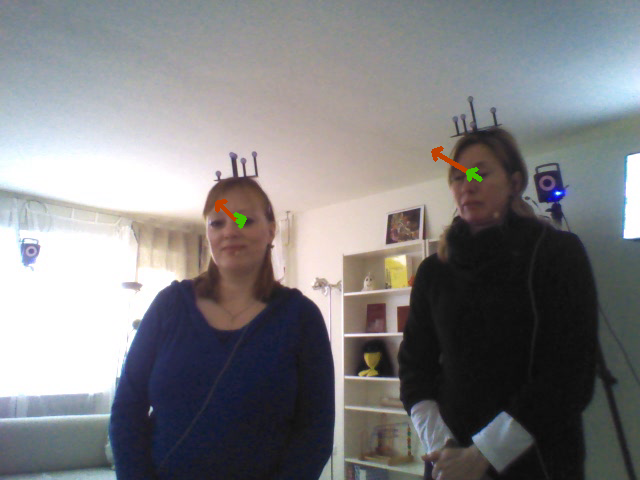}
  \includegraphics[width=.8\linewidth]{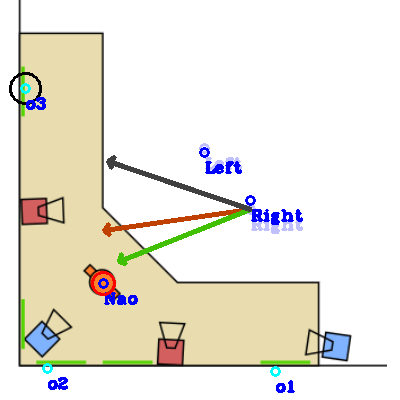}
  \includegraphics[width=.8\linewidth]{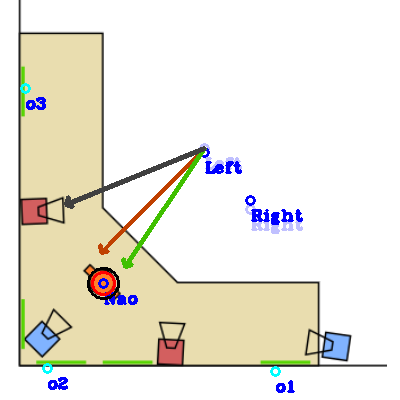}
\end{minipage}\hfill
\begin{minipage}[t]{.24\textwidth}
  \centering   \centering
  \includegraphics[width=  \linewidth]{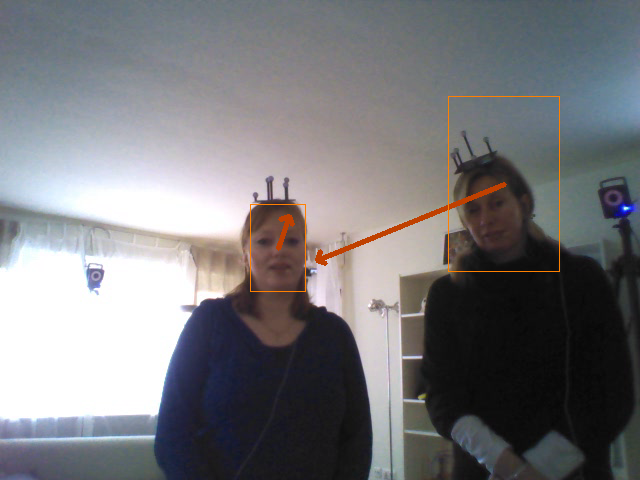}
  \includegraphics[width=.8\linewidth]{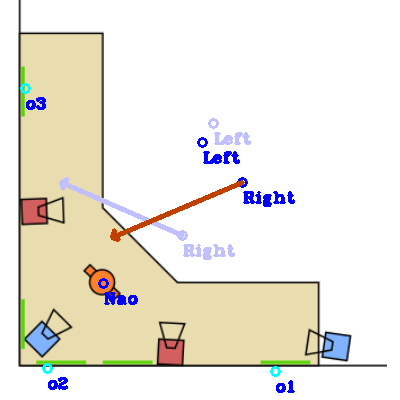}
  \includegraphics[width=.8\linewidth]{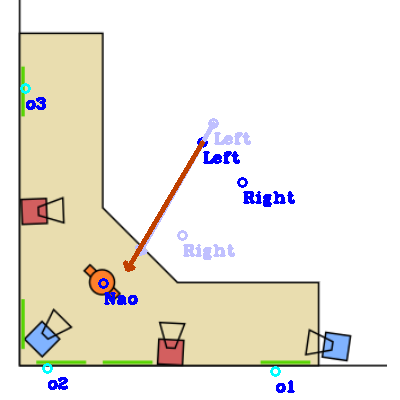}
\end{minipage}\hfill
   \caption{
     Results obtained with the proposed method on RGB data. Gaze directions are shown with green arrows, head reference directions with dark-grey arrows and observed head directions with red arrows. The ground-truth VFOA is shown with a black circle. The top row displays the image of the robot-head camera. Top views of the room show results obtained for the left person (left-p, middle row) and the right person (right-p, bottom row). %The last column shows an error in tracking that misleads the head pose estimator for the Right person.
}
\label{fig:screenshots_visual}
\end{figure*}
In the case of RGB data, the comparison between our method and the method of  \cite{Sheikhi2015} is biased by the use of different head orientation and 3D head position estimators. Indeed, the RGB data results reported in \cite{Sheikhi2015} were obtained with unpublished methods for estimating head orientations and 3D head positions, and for head tracking.
%The latter smoothes the head motion trajectories and filters out noisy measurements.
Moreover,  \cite{Sheikhi2015} uses cross-modal information, namely the speaker identity based on the audio track (one of the participants or the robot) as well as the identity of the object of interest.
We also note that \cite{Sheikhi2015} reports mean FRR values obtained over all the test recordings, instead of an FRR value for each recording.
Table~\ref{table:global_results} summarizes a comparison between the average FRR obtained with our method, with \cite{Ba2009}, and with \cite{Sheikhi2015}.  Our method yields a similar FRR score as \cite{Sheikhi2015} using the Vicon data (first row) in which case the same head pose inputs are used.
%\cite{Sheikhi2015} uses additional audio-track information that is not used by our method.

\begin{table}[t!]
  \centering
  \caption{Mean FRR scores obtained with \cite{Ba2009}, with \cite{Sheikhi2015} and with the proposed method. Recording \#26 was excluded from the FRR means as reported in  \cite{Sheikhi2015}. Moreover, \cite{Sheikhi2015} uses additional contextual information.}
  \label{table:global_results}
  \begin{tabular}{|c|c|c|c|}
    \hline
                & Ba \& Odobez~\cite{Ba2009} & Sheikhi~\cite{Sheikhi2015}&Proposed\\
    \hline
    Vicon data  & 56.5             &\textbf{66.6}              & 64.7   \\
    \hline
    RGB data & 39.0             &\textbf{62.4}              & 54.7   \\
    \hline
  \end{tabular}
\end{table}

% The differences in~\ref{table:global_results} for the baseline method is probably due, marginally to the value of $\alphabf$ used. More importantly, the difference from our version of the baseline is probably due to the visual estimation adequacy \wrt to the task.

%\subsection{Discussion}
%
%In a natural scenario such as in the \emph{Vernissage} dataset, people don't necessarily move the head synchronously with their gaze. An example is given in figure~\ref{fig:eye_only}. This actually draws limits on achievable precision of methods using only head pose to estimate gaze direction; in fact, we believe that it is not possible to consistently achieve better results without using additional information. % as \eg Sheikhi et al did in~\cite{Sheikhi2015}, using the robot contextual information.
%%
%\begin{figure}
%  \centering
%  \includegraphics[width=0.30\linewidth]{before_eye_switch} \label{subfig:look_left}
%  \includegraphics[width=0.30\linewidth]{after_eye_switch} \label{subfig:look_right}
%  \caption{Two successive frames of a person that changes gaze direction without moving the head (red arrow).}
%  \label{fig:eye_only}
%\end{figure}
%

\subsection{LAEO Dataset}
\label{subsec:laeo-expe}
%On the \emph{LAEO} dataset, the individual VFOAs are not provided. It would then require additional annotations to perform the training. Instead, we chose to train the parameters on the whole \emph{Vernissage} dataset and use the parameters directly. The main difference between the experimental setups is that there is no passive targets in \emph{LAEO}. In particular, some of the VFOA transitions will never occur. Moreover, the camera moves in some videos. This problem can be overcome by using a moving global system, linked to the camera. In this case, it is not necessary to estimate the camera displacement, contrary to the case of a robot as in \emph{Vernissage}.
    %% As already mentioned in Section~\ref{sec:laeo-dataset} above, the \emph{LAEO} annotations are different and somehow simpler than the \emph{Vernissage} annotations. Indeed, the per-shot \emph{LAEO} annotations don't allow one to infer precise per-frame VFOA annotations, \eg in frame $k$, person $i$ looks at person $j$. Moreover, when more than two persons are present in a shot, the annotations don't specify who are the persons that look at each other. For these reasons, we decided to use the model parameters that were estimated using the \emph{Vernissage} training data.
    As already mentioned in Section~\ref{sec:laeo-dataset} above, the \emph{LAEO} annotations are incomplete to estimate the person-wise VFOA at each frame. Indeed, the only VFOA-related annotation is whether two people are looking at each other during the shot. This is not sufficient to know in which frames they are actually looking at each other. Moreover, when more than two people appear in a shot, the annotations don't specify who are the people that look at each other. For these reasons, we decided to estimate the parameters using Vicon data of the whole \emph{Vernissage} dataset, \ie cross-validation.

\begin{figure*}[t!]
\centering
\begin{tabular}{cccc}
\includegraphics[height=0.135\linewidth]{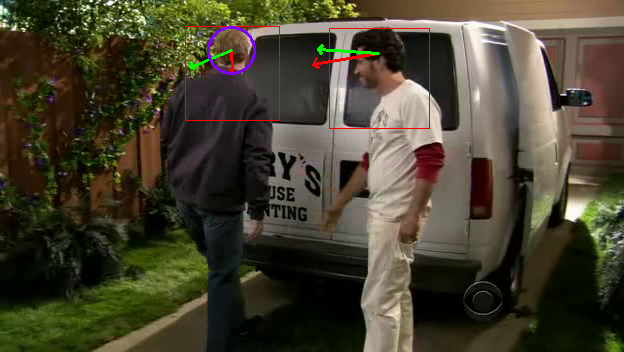} &
\includegraphics[height=0.135\linewidth]{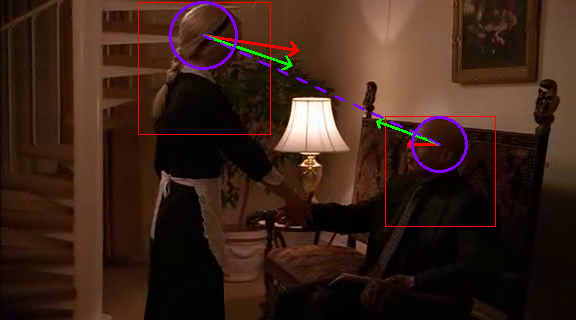} &
\includegraphics[height=0.135\linewidth]{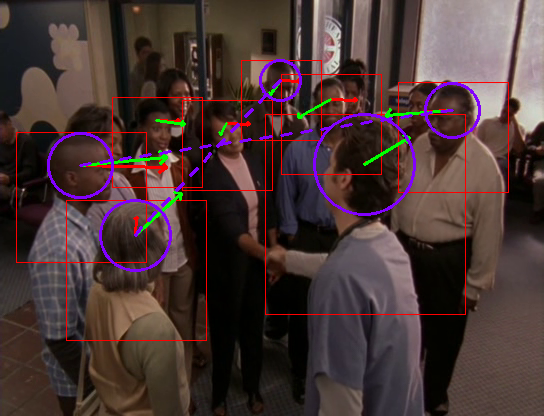} &
\includegraphics[height=0.135\linewidth]{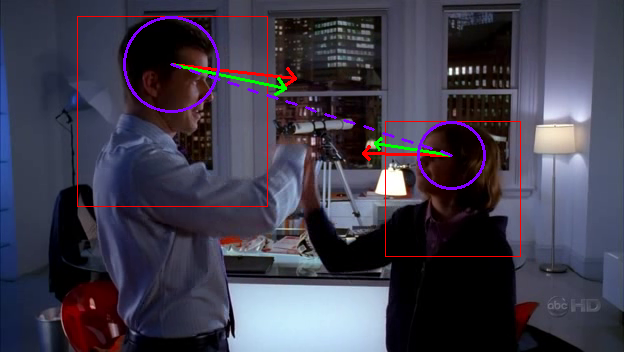} \\
\includegraphics[height=0.135\linewidth]{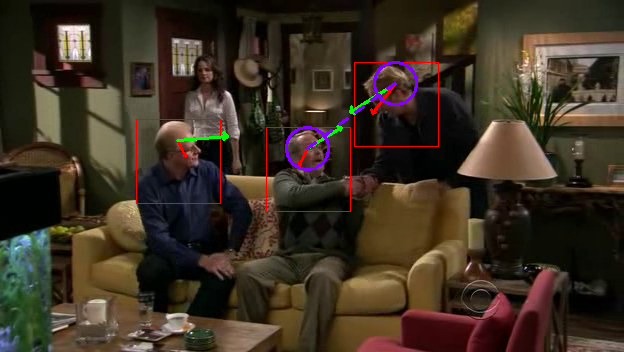} &
\includegraphics[height=0.135\linewidth]{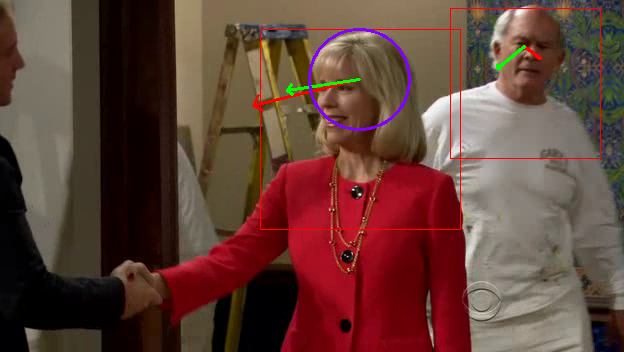} &
\includegraphics[height=0.135\linewidth]{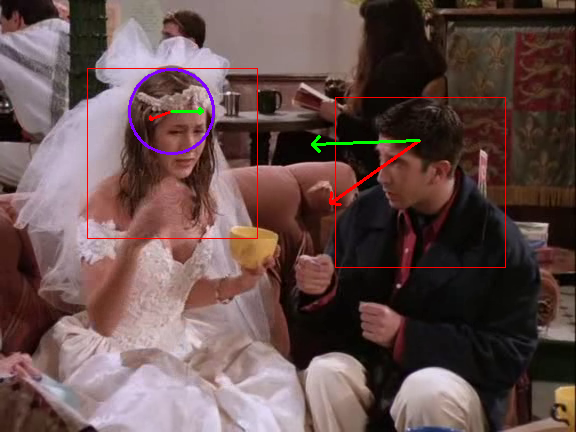} &
\includegraphics[height=0.135\linewidth]{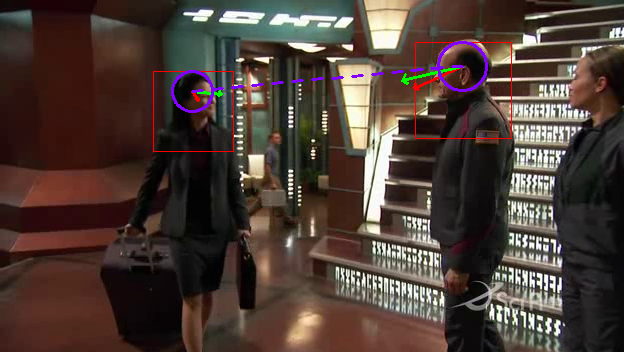} \\
\end{tabular}
\caption{\label{fig:laeo-examples} This figure shows some results obtained with the \emph{LAEO} dataset. The top row shows results obtained with coarse head orientation and the bottom row shows results obtained with fine head orientation. Head orientations are shown with red arrows. The algorithm infers gaze directions (green arrows) and VFOAs (blue circles). People looking at each others are shown with a dashed blue line.
}
\end{figure*}

We used the same pipeline as with the \emph{Vernissage} RGB data to estimate 3D head positions,  $\tilde{\X}_{1:T}$, from the face bounding boxes. Concerning head orientation, there are two cases: coarse head orientations (manually annotated) and fine head orientations (estimated). Coarse head orientations were obtained in the following way: pan and tilt values were associated with each head orientation label, namely the pan angles $-20\degree$, $20\degree$, $-80\degree$, $80\degree$, and $180\degree$ were assigned to labels \textit{frontal-left}, \textit{frontal-right}, \textit{profile-left}, \textit{profile-right}, and \textit{backwards} respectively, while a tilt anble of $0\degree$ was assigned to all five labels. Fine head orientations were estimated using the same procedure as in the case of the \emph{Vernissage} RGB data, namely face detection, face tracking, and head orientation estimation using \cite{Drouard2017}.
Algorithm~\ref{algo:inference} was used to compute the VFOA for each frame and for each person thus allowing to determine who looks at whom, \eg Fig.~\ref{fig:laeo-examples}.

We used two shot-wise, not frame-wise, metrics since the \emph{LAEO} annotations are for each shot: the \textit{shot recognition rate} (SRR), \eg Table~\ref{table:laeo-lrr}, and the \textit{average precision} (AP), \eg Table~\ref{table:laeo-map}. We note that~\cite{Marin-Jimenez2014} only provides AP scores.
It is interesting to note that the proposed method yields results comparable with those of \cite{Marin-Jimenez2014} on this dataset. This is quite remarkable knowing that we estimated the model parameters with the \emph{Vernissage} training data.

%We used the same metrics as in \cite{Marin-Jimenez2014} to quantitatively compare our method with theirs: the \textit{LAEO recognition rate} (LRR), \eg Table~{table:laeo-lrr}, and the \textit{mean average precision} (mAp), \eg Table~\ref{table:laeo-map}.
%yields the VFOAs $\V_{1:T}$. Finally, we can compute on each frame whether there are two people looking at each other or not. This result is smoothed across time to give a score over the current shot. The following table gives the proportion of shots with Looking status correctly recognized (LRR or Laeo Recognition Rate) with an appropriate threshold.

  \begin{table}[h!]
    \centering
    \caption{Average shot recognition rate (SRR) obtained with \cite{Ba2009} and with the proposed method.}
    \label{table:laeo-lrr}
    \begin{tabular}{|c|c|c|}
      \hline
      & Ba \& Odobez~\cite{Ba2009} & Proposed\\
      \hline
      Coarse head orientation & 0.535            & \textbf{0.727}\\
      \hline
      Fine head orientation   & 0.363            & \textbf{0.479}\\
      \hline
    \end{tabular}
  \end{table}

  \begin{table}[h!]
    \centering
    \caption{Average precision (AP) obtained with \cite{Marin-Jimenez2014}, with Ba \& Odobez \cite{Ba2009} and with the proposed method.}
    \label{table:laeo-map}
    \begin{tabular}{|c|c|c|c|}
      \hline
      & Marin-Jimenez et al. \cite{Marin-Jimenez2014} & \cite{Ba2009} & Proposed\\
      \hline
      Coarse head orientation & \textbf{0.925}  & 0.916   & 0.923\\
      \hline
      Fine head orientation   & \textbf{0.896}  & 0.838   & 0.890\\
      \hline
    \end{tabular}
  \end{table}

%% file: section_conclusion.tex
\section{Conclusions}

In this paper we addressed the problem of estimating and tracking gaze and visual focus of attention of a group of participants involved in social interaction. We proposed a Bayesian dynamic formulation that exploits the correlation between head movements and eye gaze on one side, and between visual focus of attention and eye gaze on the other side. We described in detail the proposed model. In particular we showed that the entries of the large-sized matrix of VFOA transition probabilities have a very small number of different possibilities for which we provided closed-form formulae. The immediate consequence of this simplified transition matrix is that the associated learning doesn't require a large training dataset. We showed that the problem of simultaneously inferring VFOAs and gaze directions over time can be cast in the framework of a switching state-space model, which yields tractable learning and inference algorithms.

We applied the proposed method to two datasets, \emph{Vernissage} and \emph{LAEO}. \emph{Vernissage} contains several recordings of a human-robot interaction scenario. We experimented both with motion capture data gathered with a Vicon system and with RGB data gathered with a camera mounted onto a robot head. We also experimented with the \emph{LAEO} dataset that contains several hundreds of video shots extracted from TV shows.
%% As with any machine learning formalisms, the effectiveness of our method depends on the quality of an annotated training dataset. In the case of the \emph{Vernissage} dataset, the VFOAs (target $j$ that is gazed by person $i$, for each person and in every frame) were manually annotated. Clearly these annotations are prone to errors because there is no quantitative way to measure the actual VFOA of a person whose eyes are not visible. This explains some of the large errors between the manually annotated VFOAs and the estimated VFOAs, both with the Vicon and the RGB data.
A quite remarkable result is that the parameters obtained by training the model with the \emph{Vernissage} data have been successfully used for testing the method with the \emph{LAEO} data, \ie cross-validation. This can be explained by the fact that social interactions, even in different contexts, share a lot of characteristics.
We compared our method with three other methods, based on HMMs \cite{Ba2009}, on input-output HMMs \cite{Sheikhi2015}, and on a geometric model \cite{Marin-Jimenez2014}.
The interest of these methods (including ours) resides in the fact that eye detection, unlike many existing gaze estimation methods, is not needed. This feature makes the above methods practical and effective in a very large number of situations, \eg social interaction.

We note that gaze inference from head orientation is an ill-posed problem.
Indeed, the correlation between gaze and head movements is person dependent as well as context dependent.
It is however important to detect gaze whenever the eyes cannot be reliably extracted from images and properly analyzed. We proposed to solve the problem based on the fact that alignments often occur between gaze directions and a finite number of targets, which is a sensible assumption in practice.

Contextual information could considerably improve the results. Indeed, additional information such as speaker recognition (as in \cite{Sheikhi2015}), speaker localization \cite{LiTASLP2017}, speech recognition, or speech-turn detection \cite{GebruPAMI2017} may be used to learn VFOA transitions in multi-party multimodal dialog systems.

In the future we plan to investigate discriminative methods based on neural network architectures for inferring gaze directions from head orientations and from contextual information. For example one could train a deep learning network from input-output pairs of head pose and visual focus of attention. For this purpose, one can combine a multiple-camera system, to accurately detect the eyes of several participants and to estimate their head poses, with a microphone-array and associated algorithms to infer speaker and speech information.

%- Estimer le gaze ˆ partir de la head pose n'est pas un problme bien posŽ. Il y aura toujours des ambigu•tŽs.
%-> On pourrait envisager d'estimer l'orientation des yeux directement ˆ partir de l'image. Ou bien d'estimer de manire jointe le gaze et la head pose ˆ partir de l'image. Ceci pourrait facilement utiliser du deep learning.
%-> On ne l'a pas encore fait car il n'existe pas de dataset sur lequel on peut s'entra"ner. Les datasets existants pour les yeux reposent sur des gens qui regardent un Žcran, et a se gŽnŽralise mal ˆ un VFOA 3D.
%
%- Online parameter learning: Le robot peut s'adapter aux personnes en face de lui (modifier les paramtres pendant l'infŽrence)
%- Reinforcement learning: Concevoir un scŽnario dans lequel le robot peut avoir un retour et sait s'il a bien devinŽ le VFOA ou non
%
%- Saliency
%-> Comprendre ˆ partir du regard des gens ce qui est important dans la scne
%
%- Quelle commande le robot doit effectuer pour mieux rŽsoudre un des problmes ci-dessus ? O regarder/tourner la tte ?
%
%
%Voila. Je te laisse prendre ce qui te semble le plus "sexy" pour la conclusion.
%En ce qui me concerne, je vais essayer d'harmoniser les citations, en comparant ˆ ce que d'autres papiers PAMI ont fait.
%Benoit

%% file: appendix_vfoa_transition_cases.tex
\section{VFOA Transition Probabilities}
\label{app:transitions_list}

% \def\Tset {\mathcal{T}}
% \def\Aset {\mathcal{A}}
% \def\Pset {\mathcal{P}}
% \def\Oset {\mathcal{O}}

% notations:
% \begin{itemize}
% \item $\Aset$ is the set of active targets $\{1\ldots N\}$
% \item $\Pset$ is the set of passive targets $\{N+1\ldots N+M\}$
% \item $\Tset$ is the set of all targets $\Aset\cup\Pset$
% \item $\Oset$ is the set of all possible VFOA $\Tset\cup\{0\}$
% \end{itemize}

Using the notations introduced in section~
%\ref{subsec:vfoa_dynamics}
III-C, 
let $i$, $1\leq i \leq N$, be an active target. In section~
%\ref{subsec:vfoa_dynamics} 
III-C
we showed that in practice the probability transition matrix has up to 15 different entries. For completeness, these entries are listed below.

\begin{itemize}
\item The VFOA of $i$ at $t-1$ is neither an active nor a passive target ($k=0$):
% \paragraph{First case : $\VV_{t-1}^i=0$}
\begin{align*}
  \pizz=&P(\VV_t^i=0|\VV_{t-1}^i=0) \\
  \pijz=&P(\VV_t^i=j|\VV_{t-1}^i=0) \\
  % &P(\VV_t^i=j|\VV_{t-1}^i=0) && \forall j\in \Tset\backslash \{i\}  \\
\end{align*}
  % 
  % \paragraph{Second case : $\VV_{t-1}^i=k\in\Pset$}
  \item The VFOA of $i$ at $t-1$ is a passive target ($N<k \leq N+M$):
\begin{align*}
  \pizk=&P(\VV_t^i=0|\VV_{t-1}^i=k) \\
  \pikk=&P(\VV_t^i=k|\VV_{t-1}^i=k) \\
  \pijk=&P(\VV_t^i=j|\VV_{t-1}^i=k) \\
  % \pizz=&P(\VV_t^i=j|\VV_{t-1}^i=k) \pizz=&\pizz=& \forall j\in \Tset\backslash \{i,k\} \\
\end{align*}
  % 
  % \paragraph{Third case : $\VV_{t-1}^i=k\in\Aset\backslash\{i\}, \VV_{t-1}^k=0$}
  \item The VFOA of $i$ at $t-1$ is an active target ($1\leq k \leq N, k\neq i$):
 \begin{align*}
  \pizkz=&P(\VV_t^i=0|\VV_{t-1}^i=k, \VV_{t-1}^k=0) \\
  \pikkz=&P(\VV_t^i=k|\VV_{t-1}^i=k, \VV_{t-1}^k=0) \\
  \pijkz=&P(\VV_t^i=j|\VV_{t-1}^i=k, \VV_{t-1}^k=0) \\
  % \pizz=&P(\VV_t^i=j|\VV_{t-1}^i=k, \VV_{t-1}^k=0) \pizz=&\pizz=& \forall j\in \Tset\backslash \{i,k\} \\
  % \end{align}
  % 
  % \paragraph{Fourth case : $\VV_{t-1}^i=k\in\Aset\backslash\{i\}, \VV_{t-1}^k=i$}
  % \begin{align}
  \pizki=&P(\VV_t^i=0|\VV_{t-1}^i=k, \VV_{t-1}^k=i) \\
  \pikki=&P(\VV_t^i=k|\VV_{t-1}^i=k, \VV_{t-1}^k=i) \\
  \pijki=&P(\VV_t^i=j|\VV_{t-1}^i=k, \VV_{t-1}^k=i) \\
  % \pizz=&P(\VV_t^i=j|\VV_{t-1}^i=k, \VV_{t-1}^k=i) \pizz=&\pizz=& \forall j\in \Tset\backslash \{i,k\} \\
  % \end{align}
  % 
  % \paragraph{Fifth case : $\VV_{t-1}^i=k\in\Aset\backslash\{i\}, \VV_{t-1}^k=l\in\Tset\backslash\{i\}$}
  % \begin{align}
  \pizkl=&P(\VV_t^i=0|\VV_{t-1}^i=k, \VV_{t-1}^k=l) \\
  \pikkl=&P(\VV_t^i=k|\VV_{t-1}^i=k, \VV_{t-1}^k=l) \\
  \pilkl=&P(\VV_t^i=l|\VV_{t-1}^i=k, \VV_{t-1}^k=l) \\
  \pijkl=&P(\VV_t^i=j|\VV_{t-1}^i=k, \VV_{t-1}^k=l)
  % \pizz=&P(\VV_t^i=j|\VV_{t-1}^i=k, \VV_{t-1}^k=l) \pizz=&\pizz=& \forall j\in \Tset\backslash \{i,k,l\}\nonumber
\end{align*}
\end{itemize}
%Some of these cases may never happen if there is not enough targets ($M=0$ and/or $N<3$)

%% file: appendix_vfoa_learning.tex
%\newpage
\section{VFOA Learning}
\label{app:vfoa-learning}

This appendix provides the formulae allowing to estimate the 15 transitions probabilities as explained in section~V-A.
%\ref{subsec:learning-vfoa-trans}.

\begin{align*}
  \pizzhat = &\frac{\displaystyle\sum_{q=1}^{Q}\sum_{i=2}^{N_q}\sum_{t=2}^{T_q}  \delta_0(\VV_t^{q,i})\delta_0(\VV_{t-1}^{q,i})}{\displaystyle\sum_{q=1}^{Q}\sum_{i=2}^{N_q}\sum_{t=2}^{T_q} \delta_0(\VV_{t-1}^{q,i})} 
  \end{align*}
\begin{align*}
  \pijzhat = &\frac{\displaystyle\sum_{q=1}^{Q}\sum_{i=2}^{N_q}\sum_{t=2}^{T_q}\sum_{j\neq i} \delta_j(\VV_t^{q,i}) \delta_0(\VV_{t-1}^{q,i})}{\displaystyle\sum_{q=1}^{Q}\sum_{i=2}^{N_q}\sum_{t=2}^{T_q} \delta_0(\VV_{t-1}^{q,i})}\nonumber%\\
\end{align*}
\begin{align*}
  \pizkhat =&\frac{\displaystyle\sum_{q=1}^{Q}\sum_{i=2}^{N_q}\sum_{t=2}^{T_q}\sum_{k=N_q+1}^{N_q+M_q} \delta_0(\VV_t^{q,i}) \delta_k(\VV_{t-1}^{q,i})}{\displaystyle\sum_{q=1}^{Q}\sum_{i=2}^{N_q}\sum_{t=2}^{T_q}\sum_{k=N_q+1}^{N_q+M_q} \delta_k(\VV_{t-1}^{q,i})} 
  \end{align*}
\begin{align*}
  \pikkhat =&\frac{\displaystyle\sum_{q=1}^{Q}\sum_{i=2}^{N_q}\sum_{t=2}^{T_q}\sum_{k=N_q+1}^{N_q+M_q} \delta_k(\VV_t^{q,i}) \delta_k(\VV_{t-1}^{q,i})}{\displaystyle\sum_{q=1}^{Q}\sum_{i=2}^{N_q}\sum_{t=2}^{T_q}\sum_{k=N_q+1}^{N_q+M_q} \delta_k(\VV_{t-1}^{q,i})} 
   \end{align*}
\begin{align*}
  \pijkhat =&\frac{\displaystyle\sum_{q=1}^{Q}\sum_{i=2}^{N_q}\sum_{t=2}^{T_q}\sum_{k=N_q+1}^{N_q+M_q}\sum_{j\neq i,k} \delta_j(\VV_t^{q,i}) \delta_k(\VV_{t-1}^{q,i})}{\displaystyle\sum_{q=1}^{Q}\sum_{i=2}^{N_q}\sum_{t=2}^{T_q}\sum_{k=N_q+1}^{N_q+M_q} \delta_k(\VV_{t-1}^{q,i})}
\end{align*}
\begin{align*}
  \pizkzhat =&\frac{\displaystyle\sum_{q=1}^{Q}\sum_{i=2}^{N_q}\sum_{t=2}^{T_q}\sum_{\substack{k=1\\k\neq i}}^{N_q} \delta_0(\VV_t^{q,i}) \delta_k(\VV_{t-1}^{q,i}) \delta_0(\VV_{t-1}^{q,k})}{\displaystyle\sum_{q=1}^{Q}\sum_{i=2}^{N_q}\sum_{t=2}^{T_q}\sum_{\substack{k=1\\k\neq i}}^{N_q} \delta_k(\VV_{t-1}^{q,i}) \delta_0(\VV_{t-1}^{q,k})} 
   \end{align*}
\begin{align*}
  \pikkzhat =&\frac{\displaystyle\sum_{q=1}^{Q}\sum_{i=2}^{N_q}\sum_{t=2}^{T_q}\sum_{\substack{k=1\\k\neq i}}^{N_q} \delta_k(\VV_t^{q,i}) \delta_k(\VV_{t-1}^{q,i}) \delta_0(\VV_{t-1}^{q,k})}{\displaystyle\sum_{q=1}^{Q}\sum_{i=2}^{N_q}\sum_{t=2}^{T_q}\sum_{\substack{k=1\\k\neq i}}^{N_q} \delta_k(\VV_{t-1}^{q,i}) \delta_0(\VV_{t-1}^{q,k})} 
   \end{align*}
\begin{align*}
  \pijkzhat =&\frac{\displaystyle\sum_{q=1}^{Q}\sum_{i=2}^{N_q}\sum_{t=2}^{T_q}\sum_{\substack{k=1\\k\neq i}}^{N_q}\sum_{j\neq i,k} \delta_j(\VV_t^{q,i})\delta_k(\VV_{t-1}^{q,i}) \delta_0(\VV_{t-1}^{q,k})}{\displaystyle\sum_{q=1}^{Q}\sum_{i=2}^{N_q}\sum_{t=2}^{T_q}\sum_{\substack{k=1\\k\neq i}}^{N_q} \delta_k(\VV_{t-1}^{q,i}) \delta_0(\VV_{t-1}^{q,k})}
\end{align*}
\begin{align*}
  \pizkihat =&\frac{\displaystyle\sum_{q=1}^{Q}\sum_{i=2}^{N_q}\sum_{t=2}^{T_q}\sum_{\substack{k=1\\k\neq i}}^{N_q} \delta_0(\VV_t^{q,i}) \delta_k(\VV_{t-1}^{q,i}) \delta_i(\VV_{t-1}^{q,k})}{\displaystyle\sum_{q=1}^{Q}\sum_{i=2}^{N_q}\sum_{t=2}^{T_q}\sum_{\substack{k=1\\k\neq i}}^{N_q} \delta_k(\VV_{t-1}^{q,i}) \delta_i(\VV_{t-1}^{q,k})} 
   \end{align*}
\begin{align*}
  \pikkihat =&\frac{\displaystyle\sum_{q=1}^{Q}\sum_{i=2}^{N_q}\sum_{t=2}^{T_q}\sum_{\substack{k=1\\k\neq i}}^{N_q} \delta_k(\VV_t^{q,i}) \delta_k(\VV_{t-1}^{q,i}) \delta_i(\VV_{t-1}^{q,k})}{\displaystyle\sum_{q=1}^{Q}\sum_{i=2}^{N_q}\sum_{t=2}^{T_q}\sum_{\substack{k=1\\k\neq i}}^{N_q} \delta_k(\VV_{t-1}^{q,i}) \delta_i(\VV_{t-1}^{q,k})} 
   \end{align*}
\begin{align*}
  \pijkihat =&\frac{\displaystyle\sum_{q=1}^{Q}\sum_{i=2}^{N_q}\sum_{t=2}^{T_q}\sum_{\substack{k=1\\k\neq i}}^{N_q}\sum_{j\neq i,k} \delta_j(\VV_t^{q,i})\delta_k(\VV_{t-1}^{q,i}) \delta_i(\VV_{t-1}^{q,k})}{\displaystyle\sum_{q=1}^{Q}\sum_{i=2}^{N_q}\sum_{t=2}^{T_q}\sum_{\substack{k=1\\k\neq i}}^{N_q} \delta_k(\VV_{t-1}^{q,i}) \delta_i(\VV_{t-1}^{q,k})}\nonumber%\\
\end{align*}
\begin{align*}
  \pizklhat =&\frac{\displaystyle\sum_{q=1}^{Q}\sum_{i=2}^{N_q}\sum_{t=2}^{T_q}\sum_{\substack{k=1\\k\neq i}}^{N_q}\sum_{l\neq i,k} \delta_0(\VV_t^{q,i}) \delta_k(\VV_{t-1}^{q,i}) \delta_l(\VV_{t-1}^{q,k})}{\displaystyle\sum_{q=1}^{Q}\sum_{i=2}^{N_q}\sum_{t=2}^{T_q}\sum_{\substack{k=1\\k\neq i}}^{N_q}\sum_{l\neq i,k} \delta_k(\VV_{t-1}^{q,i}) \delta_l(\VV_{t-1}^{q,k})} 
   \end{align*}
\begin{align*}
  \pikklhat =&\frac{\displaystyle\sum_{q=1}^{Q}\sum_{i=2}^{N_q}\sum_{t=2}^{T_q}\sum_{\substack{k=1\\k\neq i}}^{N_q}\sum_{l\neq i,k} \delta_k(\VV_t^{q,i}) \delta_k(\VV_{t-1}^{q,i}) \delta_l(\VV_{t-1}^{q,k})}{\displaystyle\sum_{q=1}^{Q}\sum_{i=2}^{N_q}\sum_{t=2}^{T_q}\sum_{\substack{k=1\\k\neq i}}^{N_q}\sum_{l\neq i,k} \delta_k(\VV_{t-1}^{q,i}) \delta_l(\VV_{t-1}^{q,k})} 
   \end{align*}
\begin{align*}
 \pilklhat =&\frac{\displaystyle\sum_{q=1}^{Q}\sum_{i=2}^{N_q}\sum_{t=2}^{T_q}\sum_{\substack{k=1\\k\neq i}}^{N_q}\sum_{l\neq i,k} \delta_l(\VV_t^{q,i}) \delta_k(\VV_{t-1}^{q,i}) \delta_l(\VV_{t-1}^{q,k})}{\displaystyle\sum_{q=1}^{Q}\sum_{i=2}^{N_q}\sum_{t=2}^{T_q}\sum_{\substack{k=1\\k\neq i}}^{N_q}\sum_{l\neq i,k} \delta_k(\VV_{t-1}^{q,i}) \delta_l(\VV_{t-1}^{q,k})} 
  \end{align*}
\begin{align*}
  \pijklhat =&\frac{\displaystyle\sum_{q=1}^{Q}\sum_{i=2}^{N_q}\sum_{t=2}^{T_q}\sum_{\substack{k=1\\k\neq i}}^{N_q}\sum_{l\neq i,k}\sum_{j\neq i,k,l} \delta_j(\VV_t^{q,i}) \delta_k(\VV_{t-1}^{q,i}) \delta_l(\VV_{t-1}^{q,k})}{\displaystyle\sum_{q=1}^{Q}\sum_{i=2}^{N_q}\sum_{t=2}^{T_q}\sum_{\substack{k=1\\k\neq i}}^{N_q}\sum_{l\neq i,k} \delta_k(\VV_{t-1}^{q,i}) \delta_l(\VV_{t-1}^{q,k})}
\end{align*}